\documentclass{article}

\usepackage[nonatbib, preprint]{neurips_2021}

\usepackage[utf8]{inputenc} % allow utf-8 input
\usepackage[T1]{fontenc}    % use 8-bit T1 fonts
\usepackage{hyperref}       % hyperlinks
\usepackage{url}            % simple URL typesetting
\usepackage{booktabs}       % professional-quality tables
\usepackage{amsfonts}       % blackboard math symbols
\usepackage{nicefrac}       % compact symbols for 1/2
\usepackage{microtype}      % microtypography
\usepackage{xcolor}         % colors

\usepackage{graphicx}
\usepackage{tikz}
\usepackage{comment}
\usepackage{amsmath,amssymb} 
\usepackage{color}
\usepackage[accsupp]{axessibility} 
\usepackage{booktabs}
\usepackage{amssymb}
\usepackage{epsfig}
\usepackage{bbding}
\usepackage{xcolor}
\usepackage{units}
\usepackage{multirow}

\newcommand{\fyq}{\textcolor{black}}
\newcommand{\todo}{\textcolor{black}}
\newcommand{\mypm}{\scriptsize$\pm$}

\usepackage{biblatex}
\title{Wave-SAN: Wavelet based Style Augmentation Network for Cross-Domain Few-Shot Learning}

\author{Yuqian Fu$^{1}$, Yu Xie$^{2}$, Yanwei Fu$^{2}$, Jingjing Chen$^{1}$, Yu-Gang Jiang$^{1}\thanks{\ indicates corresponding author}$ 
\\$^1$Shanghai Key Lab of Intelligent Information Processing, School of Computer Science, Fudan University
\\$^2$School of Data Science, Fudan University
\\ \texttt{\{fuyq20, yxie18, yanweifu, chenjingjing, ygj\}@fudan.edu.cn}
}

\begin{document}

\maketitle

\begin{abstract}
 Previous few-shot learning (FSL) works mostly are limited to natural images of general concepts and categories. These works assume very high visual similarity between the source and target classes. In contrast, the recently proposed cross-domain few-shot learning (CD-FSL) aims at transferring knowledge from general nature images of many labeled examples to novel domain-specific target categories of only a few labeled examples. The key challenge of CD-FSL lies in the huge data shift between source and target domains, which is typically in the form of totally different visual styles. This makes it very nontrivial to directly extend the classical FSL methods to address the CD-FSL task. To this end, this paper studies the problem of CD-FSL by spanning the style distributions of the source dataset. Particularly, wavelet transform is introduced to enable the decomposition of visual representations into low-frequency components such as shape and style and high-frequency components e.g., texture. To make our model robust to visual styles, the source images are augmented by swapping the styles of their low-frequency components with each other. We propose a novel Style Augmentation (StyleAug) module to implement this idea. Furthermore, we present a Self-Supervised Learning (SSL) module to ensure the predictions of style-augmented images are semantically similar to the unchanged ones. This avoids the potential semantic drift problem in exchanging the styles. Extensive experiments on two CD-FSL benchmarks show the effectiveness of our method. Our codes and models will be released.
\end{abstract}

%%%%%%%%% BODY TEXT
\section{Introduction}\label{sec:intro}

% p1: FSL 
Deep learning has achieved remarkable success in many visual tasks while training a deep neural network still relies on a large amount of manually annotated data. To this end, few-shot learning (FSL)~\cite{snell2017prototypical,vinyals2016matching,sung2018learning,garcia2017few,li2020adversarial}, which aims at endowing the model with the ability to learn from a few labeled examples, has attracted numerous research interests in recent years.

% p2: CD-FSL
Generally, existing algorithms simplify the FSL problem by assuming that the target data come from the same domain as the source data. Unfortunately, this does not always hold in real-world applications. Consequently, this inspires the recent study on cross-domain few-shot learning (CD-FSL), which aims at learning FSL models across different visual domains. Typically, the CD-FSL transfers the knowledge from general nature images of many labeled examples to novel domain-specific target categories of only a few labeled examples.

\begin{figure}[t]
	\centering
	\includegraphics[width=0.95\linewidth]{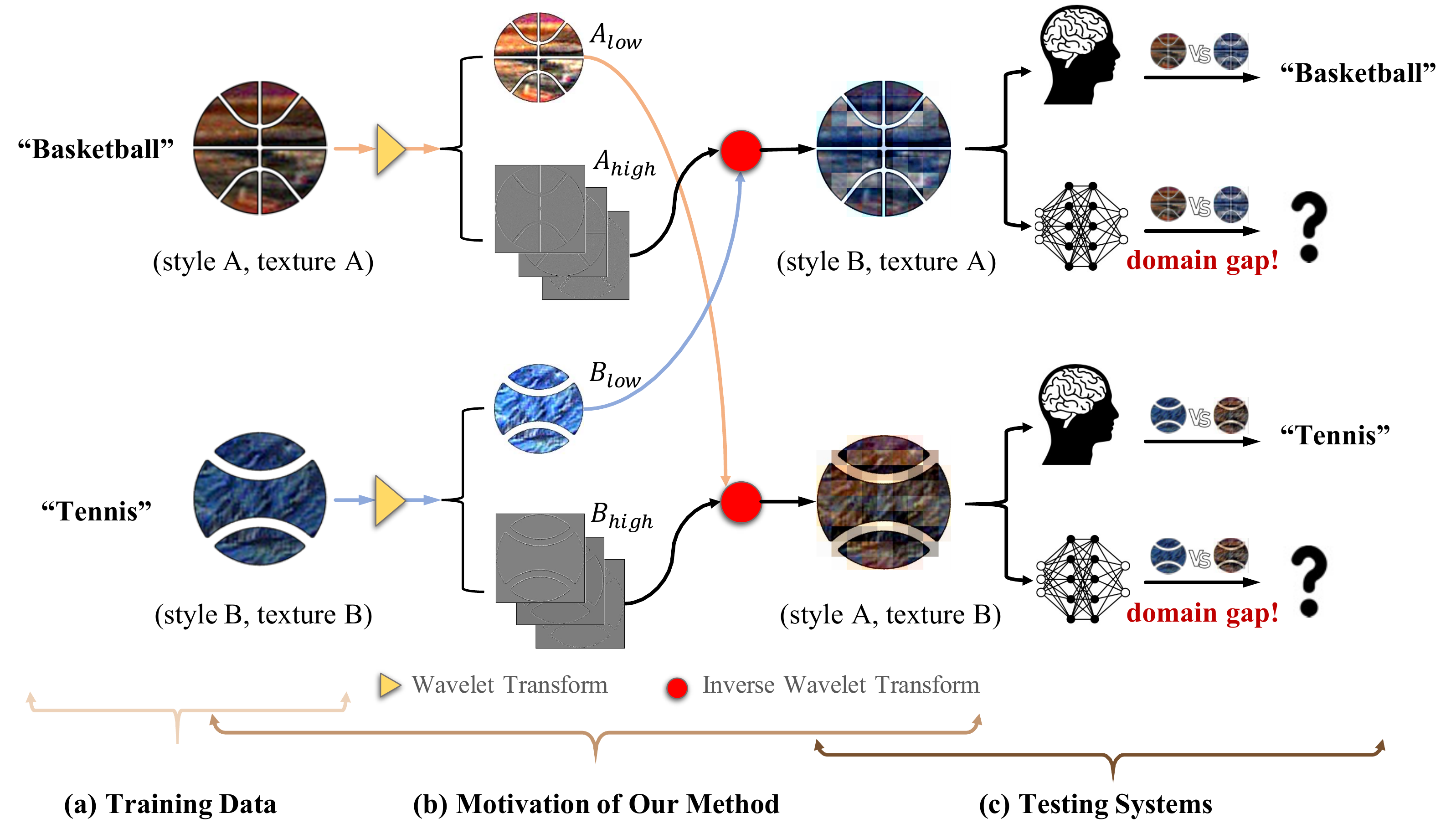}
	%\vspace{-0.1in}
	\caption{\textbf{The illustration of our motivations.} 
	\fyq{1) Given the training data with limited style as shown in Figure(a), Figure(c) indicates that when the visual appearances of the same categories change a lot, the human vision system is still able to recognize the correct categories while models are likely to fail due to the domain gap issue.}
	2) Figure(b) shows that the wavelet transform decomposes the low-frequency (shape and style) and high-frequency (texture) components. When the shapes of two input images are the same, the styles of these images can be transferred by simply exchanging the low-frequency components.}
	\label{fig:motivation} 
	\vspace{-0.25in}
\end{figure}

%p3: motivation of our method. 
This paper studies the extremely challenging CD-FSL task, where only a single source domain is provided for training. Critically, one key challenge comes from the huge data shift and visual difference between source and novel target domains, which are typically in the form of totally different visual styles. This makes it nontrivial to directly extend the classical FSL methods to address the CD-FSL tasks~\cite{chen2019closer}. 
In particular, a model trained on a single source domain of limited style diversity is likely to fail to recognize the novel target images with various visual styles changes. In contrast, the vision systems of humans are robust to different visual styles. 
For example, Figure~\ref{fig:motivation}(a) shows the training data with a limited style of ``basketball'' and ``tennis''. When it comes to recognizing the same class categories with quite different visual appearances, we humans can still classify the target images correctly by leveraging prior knowledge while the models will likely fail due to domain shift issue as in Figure~\ref{fig:motivation}(c).
To mimic such ability of our biological vision systems, it is necessary to encourage the CD-FSL models to be insensitive to visual styles. 
Inspired by such observations, our first insight
is to solve CD-FSL from a novel perspective -- \emph{spanning the distributions of the source domain by augmenting the styles of the source images.}
Critically, such an idea has less been explored in previous CD-FSL works~\cite{tseng2020cross,guo2020broader,phoo2020self,fu2021meta,sun2021explanation,guan2020large,liang2021boosting,islam2021dynamic,zhang2021shallow}.

%p4:  the introduction of our method
To this end, we propose to utilize wavelet transform to decompose the low-frequency components such as shape and style and high-frequency components e.g., texture. 
As illustrated in Figure~\ref{fig:motivation}(b), the images of the ``basketball'' and  ``tennis'' share the same shape while having different styles and textures.
After applying the wavelet transform, the shape and style are reflected in low-frequency components, while the texture is conveyed from the high-frequency components. 
Our second insight is that when the shape of two input images is the same, the styles of these two images can be well transferred by simply exchanging their low-frequency components. This inspires us to augment the styles by modifying the low-frequency components. 
Unfortunately, it is generally impractical to directly swap the low-frequency components of two images, as our source images have various complex shapes. 
Alternatively, we present a novel style augmentation (StyleAug) module by reformulating the adaptive instance normalization (AdaIN)~\cite{huang2017arbitrary}. Particularly, our StyleAug module utilizes an AdaIN based operation to exchange the styles of two low-frequency components. \todo{Besides, a recent study~\cite{geirhos2018imagenet} suggests that texture plays a critical role in image classification, 
our method thus maintains the high-frequency components unchanged to preserve the texture information.}
The style-augmented low-frequency components and the initial decomposed high-frequency components can reconstruct the augmented overall feature by applying the inverse wavelet transform.
Furthermore, augmenting styles will inevitably incur the problem of semantic drift. Thus, we present a novel self-supervised learning (SSL) module to ensure the predictions of the style-augmented images are semantically similar to the unchanged ones.

% p5: formal introduce & contribution
Formally, in this paper, a novel \textbf{wave}let based \textbf{S}tyle \textbf{A}ugmentation \textbf{N}etwork (\textbf{wave-SAN}) is proposed for CD-FSL. Our model is composed of two basic modules -- feature extractor and FSL classifier, and four novel modules -- wavelet transform, inverse wavelet transform, StyleAug module, and SSL module. We employ the meta-learning mechanism which has been the recipe for FSL and randomly sample two episodes to perform the style augmentation. Concretely, the styles of these two episodes are exchanged by inserting our wavelet transforms and StyleAug module into the feature extractor. Then, both the style-augmented episodes and the unchanged episodes are fed into the FSL classifier to obtain their predictions and corresponding FSL losses. Finally, the FSL classification losses and the consistency loss calculated by our SSL module together help optimize our model.

\noindent\textbf{Contributions.} We summarize our main contributions: 1) We propose to address the domain shift problem in CD-FSL by spanning the distributions of source styles. 2) The wavelet transform is introduced to decompose the low-frequency and high-frequency components. To the best of our knowledge, it is the first time that wavelet is used for the FSL, especially the CD-FSL task. The experimental results show that augmenting styles on the low-frequency components is better than on the overall visual feature. 3) Finally, all the key modules in our wave-SAN -- wavelet transforms, StyleAug module, and SSL module do not need any learnable parameters, thus making our wave-SAN complementary to other CD-FSL methods via a plug-and-play manner.

%%%%%%%%% Related Work
\section{Related Work}
\noindent \textbf{Few-Shot Learning.}
There are roughly three types of FSL methods: 1) metric-learning based methods~\cite{snell2017prototypical,vinyals2016matching,sung2018learning,garcia2017few} comparing the similarity between support and query images; 2) optimization-based methods~\cite{finn2017model,munkhdalai2017meta,ravi2016optimization,rusu2018meta,lee2019meta} learning to finetune the trained model with few examples; and 3) data augmentation based methods~\cite{chen2019image,li2020adversarial,hariharan2017low} learning to augment images. Generally, these methods assume that the source and target data come from the same domain, thus the performance is degraded in cross-domain FSL setting~\cite{chen2019closer}. 

\noindent \textbf{Cross-Domain Few-Shot Learning.} 
Several efforts have been made to address the CD-FSL task recently. FWT~\cite{tseng2020cross}, BSCD-FSL~\cite{guo2020broader}, STARTUP~\cite{phoo2020self}, and Meta-FDMixup~\cite{fu2021meta} mainly aim at thoroughly defining the benchmarks and settings for CD-FSL. As a result, learning CD-FSL under single/multiple source domain and with no/unlabeled/labeled target data settings are proposed. Besides, semantic labels are introduced by TriAE~\cite{guan2020large} to construct a joint feature space. 
Some other tentative attempts include~\cite{sun2021explanation,adler2020cross,das2021importance,liang2021boosting,liu2020feature,yeh2020large,wang2021cross,islam2021dynamic,cai2021damsl,zhang2021shallow,chen2021self}. Among them, LRP~\cite{sun2021explanation} uses the explanation results generated by ~\cite{bach2015pixel} to guide the learning process. ATA~\cite{wang2021cross} mainly constructs more challenging training episodes via adversarial task augmentation. ConFT~\cite{das2021importance} and NSAE~\cite{liang2021boosting} tackle the CD-FSL by utilizing contrastive loss and enhancing noises, respectively. 
Since ConFT and NSAE need to be fine-tuned on the target images, they both require additional computational cost and inference time. 
In contrast, our wave-SAN tackles CD-FSL by spanning distributions of source styles, which is directly used for inference without any fine-tuning.

\noindent \textbf{Data Augmentation.}
Vanilla methods include random cropping, rotation, flipping, and color jittering~\cite{krizhevsky2012imagenet,zeiler2014visualizing,chen2020simple}. Recently, many mixup-based algorithms~\cite{zhang2017mixup,verma2019manifold,yun2019cutmix} that mix the images and the labels simultaneously have emerged. Different from these methods, we augment the styles while keeping the semantic label unchanged. MixStyle~\cite{zhou2021domain} which mixes styles for domain adaptation is the most similar work to us. However, our wave-SAN augments the styles upon the low-frequency components rather than overall visual features.

\noindent\textbf{Deep Learning in Frequency Domains.} 
It has been widely explored in various applications 
including superresolution~\cite{bae2017beyond,guo2017deep,liu2018multi,deng2019wavelet}, deblurring~\cite{zou2021sdwnet,min2018blind,gao2019dynamic}, denoising~\cite{kang2018deep}, compression~\cite{levinskis2013convolutional,gueguen2018faster}, classification~\cite{fujieda2017wavelet,ryu2018dft,williams2018wavelet},  semantic segmentation~\cite{yang2020fda,liu2020remove},  and style transfer~\cite{yoo2019photorealistic,jamadandi2019exemplar,ding2022deep,singh2021safin}. These methods typically utilize the frequency information extracted by wavelet~\cite{wu2012eulerian,zhang1992wavelet} and Fourier transform~\cite{bracewell1986fourier}, and then learn to reconstruct the signals. 
Recently, frequency-based methods have been utilized for style transfer~\cite{yoo2019photorealistic}, and image segmentation~\cite{yang2020fda}. Different from these works, we introduce the wavelet to bridge the domain gap of CD-FSL. Particularly,  the wavelet serves as the tool to decompose the low-frequency and high-frequency components from all features in our wave-SAN. 
The key innovation of our method lies in augmenting the style distributions of source data by swapping the “styles” of the low-frequency components thus \textit{narrowing the domain gap in CD-FSL}. Besides, our purpose of maintaining the high-frequency components unchanged is to keep the texture from being destroyed for a better classification ability which has less been considered in previous attempts. %Essentially, we highlight this is the first time of employing wavelet for CD-FSL, rather than presenting a novel deep wavelet transformation module.

% \fyq{Notably, WCT$^2$~\cite{yoo2019photorealistic} seems the most related to us. However, we highlight that though it tackles the style transfer task, the wavelet is still used for better image reconstruction solely. More specifically, the wavelet/inverse wavelet transforms are used to replace the original max-pooling and unpooling layers while the style transfer work is done by its base model ~\cite{li2018closed}. By contrast, in this paper, our key innovation is to augment the style distributions of the source data by swapping the ``styles'' of the low-frequency components thus narrowing the domain gap issue for CD-FSL. In this process, the wavelet serves as the tool to decompose the low-frequency components from the overall features.}

%%%%%%%%% method
\section{Methodology}

\noindent\textbf{Problem Setup.} 
% CD-FSL task
Given  source  $D_{s} = \left\{I_{i}, y_{i}\right\}, y_{i} \in C_{s}$ and a novel target dataset $D_{t} = \left\{I_{i}, y_{i}\right\}, y_{i} \in C_{t}$, $C_{s} \cap C_{t} = \emptyset$. The task of CD-FSL supposes that $D_s$ and  $D_t$ come from different domains. The source classes in $C_{s}$ have sufficient labeled images, while  the target classes $C_{t}$ have few labeled images per class. The task is to learn a model from $D_{s}$ which can be well generalized to the class $C_{t}$ in $D_{t}$ with only few labeled instances.

Our model is trained via an episode-based strategy. To simulate the scenarios of few-shot instances in testing, meta-learning randomly samples an episode. An episode consists of a support set $S$ and a query set $Q$. Our goal is to classify the labels of query images according to the support set. In an $n$-way-$k$-shot problem, we randomly sample $n$ classes. For each selected class, $k$ labeled images are randomly sampled to construct the support set and another $q$ images are randomly sampled as queries. Formally, $S=\left\{I_i, y_i\right\}_{i=1}^{n\times k}$, $Q=\left\{I_i, y_i\right\}_{i=1}^{n\times q }$.

\begin{figure*}[t!]
		\centering
		%\vspace{-0.1in}
		{\includegraphics[width=0.95\linewidth]{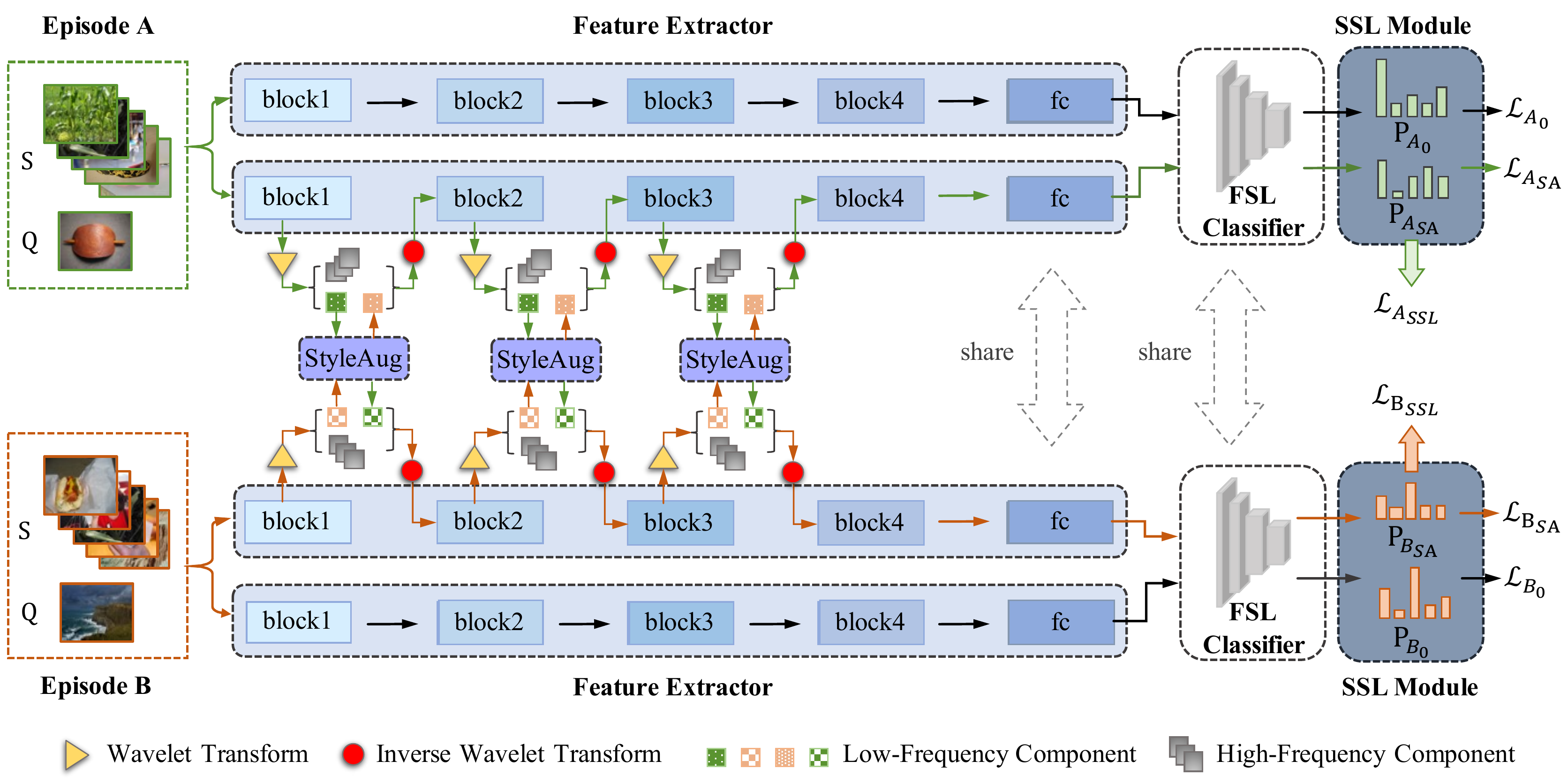}}
		\vspace{-0.1in}
		\caption{\textbf{Overview of our wave-SAN network.} In the network, a dual-episode sampling mechanism is adopted. For each episode, two predictions are generated. One is obtained by directly feeding the episode into the feature extractor and FSL classifier (black arrows), while the other one is obtained by taking the style-augmented features as input (green or orange arrows). The style augmentation is implemented by the proposed StyleAug module which exchanges the styles of two low-frequency components decomposed by wavelet transform. Then an SSL module is introduced for keeping these two predictions similar.}
		\label{fig:framework} %framework,fig1
		\vspace{-0.15in}
	\end{figure*}

	% overall & two-episode sampling 
	\noindent\textbf{Method Overview}. The overall of our wavelet based style augmentation network (wave-SAN) is summarized in Figure~\ref{fig:framework}. Generally, a dual-episode sampling method is adopted to better explore the dataset diversity of the source dataset. Particularly, for each meta-training step, we randomly sample two episodes $A$ and $B$ from the $D_s$ as the input. In particular, the forward process for these two episodes is completely symmetric, and all the modules with parameters are shared. Specifically,
	% components
	our wave-SAN is mainly composed of six components: feature extractor $\mathcal{F}_\theta$, FSL classifier $\mathcal{G}_\phi$, Discrete Wavelet Transform (DWT) layer, Inverse Wavelet Transform (IDWT) layer, Style Augmentation (StyleAug) module $\mathcal{M}_{SA}$, and Self-Supervised Learning (SSL) module $\mathcal{M}_{SSL}$, where $\theta$ and $\phi$ indicate the learnable parameters.  More concretely, feature extractor $\mathcal{F}_\theta$  extracts features $F$ from the episode input. FSL classifier $\mathcal{G}_\phi$ predicts the probabilities $P$ that the query images belong to the $n$ support categories. Note that $P \in \mathcal{R}^{nq\times n}$, again, $n$ is the number of categories in the episode, and $nq$ denotes the number of query images contained in the query set. Wavelet transform layer DWT decomposes the overall feature $F$ into a low-frequency component $F_{low}$ and high-frequency components $F_{high}$. Correspondingly, the inverse wavelet transform takes the $F_{low}$ and $F_{high}$ as input and recovers the overall feature $F$. 
	Our StyleAug module $\mathcal{M}_{SA}$ is designed to augment the source domain by exchanging the styles of two episodes, thus making the model insensitive to various styles. SSL module $\mathcal{M}_{SSL}$ forces the probabilities of augmented episode $P_{SA}$ to be the same with the initial prediction $P_{0}$ by using a Kullback-Leibler (KL) divergence loss as the regularization for consistency. It is worth mentioning that all the modules proposed by us including DWT, IDWT, $\mathcal{M}_{SA}$, and $\mathcal{M}_{SSL}$ do not introduce any additional learnable parameters; thus our method is complementary to other methods via a plug-and-play manner.
	
% overall forward pipeline
\noindent\textbf{Augmentation Strategy}.
Here we use episode $A$ to help introduce our pipeline. Given $A$, we feed it into the $\mathcal{F_\theta}$ and $\mathcal{G_\phi}$ to obtain its predictions $P_{A_{0}}$. This process is shown by the black arrows in Figure~\ref{fig:framework}. Apart from the $P_{A_{0}}$ that takes the initial episode features as the input, to improve the generalization of our model, we augment the styles of the episode and force the model to learn under such circumstances. Concretely, for the overall features $F_{A}$ extracted by the blocks of $\mathcal{F_\theta}$, e.g., block1, we first apply DWT to extract the low-frequency component $F_{A_{low}}$ and high-frequency components $F_{A_{high}}$. Then, $\mathcal{M}_{SA}$ is used to augment the styles of the $F_{A_{low}}$ by replacing the original styles of $F_{A_{low}}$ by that of $F_{B_{low}}$, generating the style-augmented low-frequency component $\Tilde{F}_{A_{low}}$. After that, the $\Tilde{F}_{A_{low}}$ and the $F_{A_{high}}$ are used to recover the style-augmented overall feature $\Tilde{F}_A$ via the IDWT. The $\Tilde{F}_A$ is further fed into the subsequent block. This process is repeated until the final feature representation is obtained. Then, $\mathcal{G_\phi}$ is used to generate the style-augmented predictions $P_{A_{SA}}$. This process is shown by the green arrows.

% losses
 \noindent\textbf{Loss Functions}.
By calculating  cross entropy  between  $P_{A_{0}}$ and $P_{A_{SA}}$ against  ground-truth, we have the loss $\mathcal{L}_{A_{0}}$ and $\mathcal{L}_{A_{SA}}$.
Besides, though styles are augmented, we expect the predictions on  semantic labels are not effected. To this end, $\mathcal{M}_{SSL}$ forces the predictions $P_{A_{SA}}$ and $P_{A_{0}}$ are similar, generating an SSL consistency loss $\mathcal{L}_{A_{SSL}}$. The same operations are applied for episode $B$, resulting in other two FSL losses $\mathcal{L}_{B_{0}}$,  $\mathcal{L}_{B_{SA}}$ and one consistency loss $\mathcal{L}_{B_{SSL}}$. These losses are used to train our model.

\subsection{Network Modules}
\begin{figure}[h]
		\centering
		\vspace{-0.15in}
		{\includegraphics[width=0.6\linewidth]{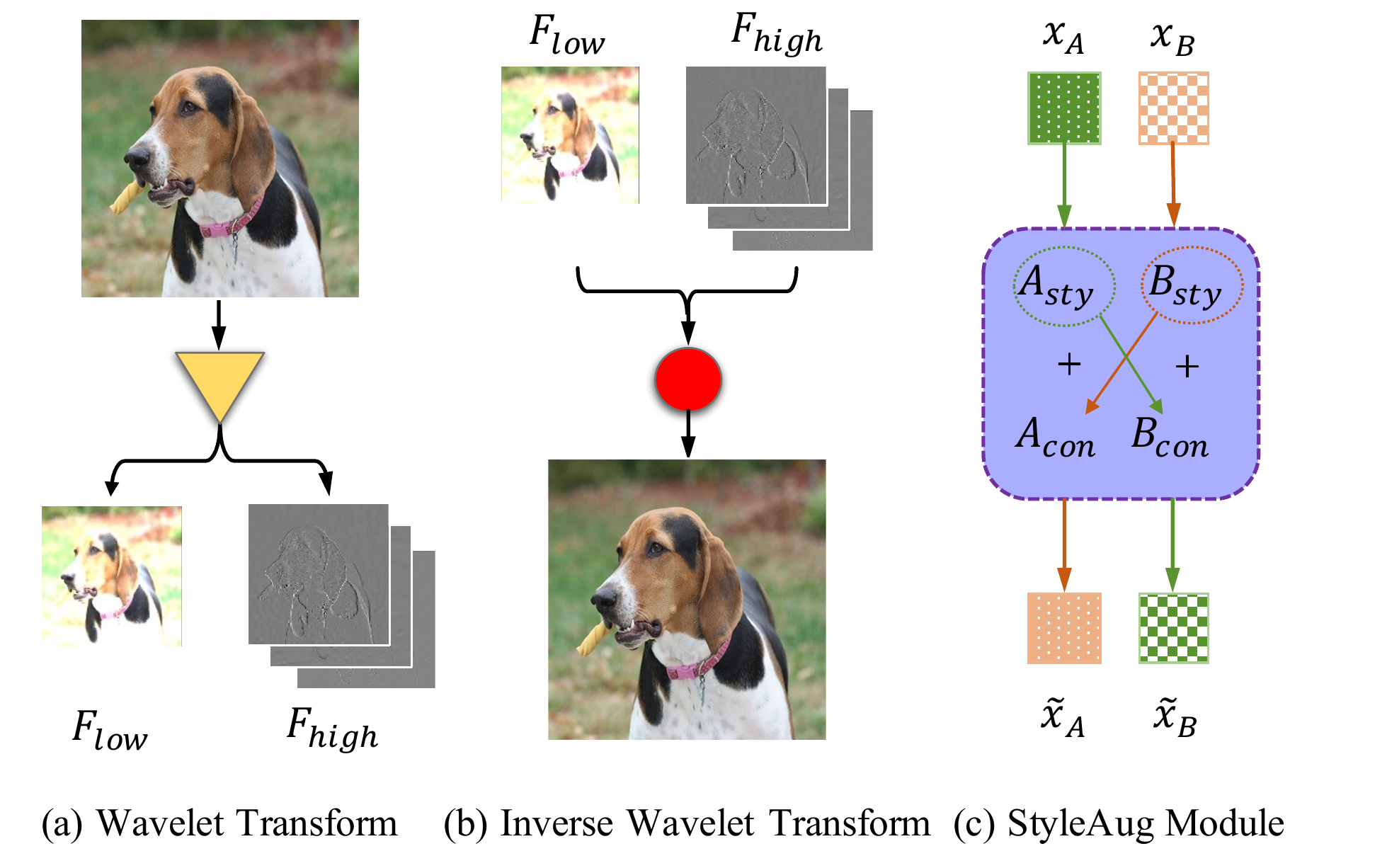}}
		\vspace{-0.15in}
		\caption{\textbf{Modules of our method.} (a) Wavelet Transform, (b) Inverse Wavelet Transform, and (c) StyleAug Module. }
		\label{fig:modules} 
		\vspace{-0.15in}
	\end{figure}

% wavelet transform
\noindent\textbf{Wavelet Transforms:} Totally two wavelet transform layers are introduced into our model. The illustration of the wavelet transform DWT is shown in Figure~\ref{fig:modules}(a). DWT decomposes the input feature map $F$ into a low-frequency component $F_{low}$ and three high-frequency components $F_{high}$. The size of the feature map $F_{low}$ and $F_{high}$ is 1/2 of $F$. Generally, $F_{high}$ conveys the texture of the input images. In this example, the edges of the dog can be clearly observed. 
While the shape and style are mainly maintained in the low-frequency component $F_{low}$. 

The illustration of the inverse wavelet transform IDWT is given in Figure~\ref{fig:modules}(b). Given low-frequency component $F_{low}$ and high-frequency components $F_{high}$, IDWT is able to reconstruct the original input $F$. Notably, DWT and the IDWT transforms are strictly invertible. We benefit from this characteristic in three ways: 1) The transforms do not lose any information; 2) The decomposition of $F_{low}$ and $F_{high}$ makes us enable to handle them separately; 3) Since the size of the input and output are the same, we can perform plug and play well.

Note that in the proposed framework, any algorithms which can serve as the tool for decomposing and reconstructing the high-frequency and low-frequency components can be used. In this paper, the most classical Haar algorithm thus is introduced as our wavelet transforms $DWT$ and $IDWT$. The details of how the Haar works are given in the supplementary materials.

	%\vspace{-0.15in}

\noindent\textbf{StyleAug Module:} Our $\mathcal{M}_{SA}$ is an  AdaIN~\cite{huang2017arbitrary} based module.  Given a content $x_{a}$ and a style image $x_{b}$ ($x_{a},x_{b}\in\mathcal{R}^{B\times C\times H\times W}$), it transfers the style of $x_{a}$ towards that of $x_{b}$ by aligning mean and variance of $x_{a}$ to those of $x_{b}$:   
	\begin{equation}
	\mathrm{AdaIn}(x_{a},x_{b})=\sigma(x_{b})\frac{x_{a}-\mu(x_{a})}{\sigma(x_{a})}+\mu(x_{b}),
	\end{equation}
	where $\mu(x)$ and $\sigma(x)$ stand for the mean and variance
	of the input $x$, respectively. Both of them are computed across the spatial dimensions:
	\begin{equation}\label{eq:mu}
	\mathrm{\mu_{b,c}}(x)=\frac{1}{HW}\sum_{h=1}^{H}\sum_{w=1}^{W}x_{b,c,h,w}
	\end{equation}
	\begin{equation}\label{eq:sigma}
	\mathrm{\sigma_{b,c}}(x)=\sqrt{\frac{1}{HW}\sum_{h=1}^{H}\sum_{w=1}^{W}(x_{b,c,h,w}-\mathrm{\mu_{b,c}}(x))^{2}+\epsilon}.
	\end{equation}

% main idea
Since a large part of the domain gap is due to \fyq{style shifts}, we propose to augment the source domain by exchanging the styles of two episodes. By virtue of such a way, the style diversity of the source dataset is well utilized to augment the initial source domain, improving the generalization ability of the model.

Our $\mathcal{M}_{SA}$ is illustrated in Figure~\ref{fig:modules}(c). Formally, given two feature maps $x_{A}$ and $x_{B}$, $\mathcal{M}_{SA}$ generates the style-augmented feature maps $\Tilde{x}_{A}$ and $\Tilde{x}_{B}$ as, 
	\begin{equation}
	 \Tilde{x}_{A} =\sigma(x_{B})\frac{x_{A}-\mu(x_{A})}{\sigma(x_{A})}+\mu(x_{B})\label{style-1}
	\end{equation}
	\begin{equation}
	 \Tilde{x}_{B} =\sigma(x_{A})\frac{x_{B}-\mu(x_{B})}{\sigma(x_{B})}+\mu(x_{A})
	\end{equation}
	
The $\mu()$ and $\sigma()$ are defined in Eq.~\ref{eq:mu} and Eq.~\ref{eq:sigma}, respectively. We augment styles upon  low-frequency components, i.e., $x_{A} = F_{A_{low}}$, $x_{B} = F_{B_{low}}$, resulting in the style-augment low-frequency components $\Tilde{F}_{A_{low}}$ and $\Tilde{F}_{B_{low}}$.

\noindent \textbf{SSL Module:} 
To eliminate the potential \fyq{semantic drift problem} caused by the StyleAug module, 
an SSL module is introduced here. Since we expect that augmenting styles will not change the semantic meanings of images, the prediction of the style-augmented episode $P_{SA}$ and the initial prediction $P_{0}$ should be principally similar with each other.

The KL divergence loss is introduced as our consistency loss. We first obtain the average probability $P_{avg}$ of these two predictions, and then we calculate the KL losses $KL(P_{SA}, P_{avg})$ and $KL(P_{0}, P_{avg})$. Finally, the average of these two KL losses is used as the final consistency loss $\mathcal{L}_{SSL}$. Formally, $\mathcal{L}_{SSL}$ is defined as:
	%\begin{equation}
	%P_{avg} = \frac{1}{2}(P_{SA} + P_{0})
	%\end{equation}
	\begin{equation}
	\mathcal{L}_{SSL}=\frac{1}{2}(KL(P_{SA},P_{avg}) + KL(P_{0},P_{avg}))
	\end{equation}
	\begin{equation}
	KL(P^1,P^2)=\frac{1}{Bn}\sum_{i=1}^{B}\sum_{j=1}^{n}P^{1}_{i,j}(\log P^{1}_{i,j}-\log P^{2}_{i,j})
	\end{equation}
	where $P^1, P^2\in\mathcal{R}^{nq\times n}, B=nq$. Empirically, comparing with calculating the KL loss between $P_{SA}$ and ${P_{0}}$ directly, our manner is easier for the model to converge.
	By using $\mathcal{M}_{SSL}$, we obtain consistency losses $\mathcal{L}_{A_{SSL}}$ and $\mathcal{L}_{B_{SSL}}$.

\subsection{Network Training and Inference}

\noindent\textbf{Network Training:} The loss function is finally composed of two initial FSL losses $\mathcal{L}_{A_{0}}$, $\mathcal{L}_{B_{0}}$, two style-augmented FSL losses $\mathcal{L}_{A_{SA}}$, $\mathcal{L}_{B_{SA}}$, and two SSL consistency loss $\mathcal{L}_{A_{SSL}}$, $\mathcal{L}_{B_{SSL}}$. The final loss function $\mathcal{L}$ is defined as follows, and the $k_1$ and $k_2$ are two hyperparameters. 
	%\begin{equation}
	%\mathcal{L}_{A}=k_{1}\mathcal{L}_{A_{0}}+k_{2}\mathcal{L}_{A_{SA}} + \mathcal{L}_{A_{SSL}}
	%\end{equation}
	%\begin{equation}
	%\mathcal{L}_{B}=k_{1}\mathcal{L}_{B_{0}}+k_{2}\mathcal{L}_{B_{SA}} + \mathcal{L}_{B_{SSL}}
	%\end{equation}
	%\begin{equation}
	%\mathcal{L}=\frac{1}{2}(\mathcal{L}_{A}+\mathcal{L}_{B})
	%\end{equation}
	\begin{equation}
	\mathcal{L}=\frac{1}{2}	\left[k_{1}(\mathcal{L}_{A_{0}} + \mathcal{L}_{B_{0}}) +k_{2}(\mathcal{L}_{A_{SA}} + \mathcal{L}_{B_{SA}})+ \mathcal{L}_{A_{SSL}} + \mathcal{L}_{B_{SSL}}\right]
	\end{equation}

\noindent \textbf{Network Testing:} The proposed DWT, IDWT, $\mathcal{M}_{SA}$, and $\mathcal{M}_{SSL}$ are employed to help improve the generalization ability of models during meta-training phase. In fact, all these four modules are directly discarded during testing phase. Thus, our testing process is very simple. For each episode, we simply feed the testing episode into the feature extractor $\mathcal{F_{\theta}}$ and the FSL classifier $\mathcal{G_{\phi}}$ to obtain its prediction. The class with the highest probability will be the predicted label.

%%%%%%%%% experiments
\section{Experiments}
% \subsection{Setup}
\noindent\textbf{Datasets.} Totally two CD-FSL benchmarks are used in our experiments. For convenience, we denote the benchmark proposed in FWT~\cite{tseng2020cross} and BSCD-FSL~\cite{guo2020broader} as FWT's benchmark and BSCD-FSL benchmark, respectively. Both of them contain five datasets and use the mini-Imagenet~\cite{deng2009imagenet} as the source dataset. For FWT's benchmark, CUB~\cite{wah2011caltech}, Cars~\cite{krause20133d}, Places~\cite{zhou2017places}, and Plantae~\cite{van2018inaturalist} serve as the target datasets, respectively. For BSCD-FSL benchmark, ChestX~\cite{wang2017chestx}, ISIC~\cite{tschandl2018ham10000,codella2019skin}, EuroSAT~\cite{helber2019eurosat}, and CropDisease~\cite{mohanty2016using} are selected as the target datasets. We follow the splits provided by FWT and BSCD-FSL.

\noindent\textbf{Network Components.} 
%We adopt the classical Haar algorithm for wavelet transforms. 
To make a fair comparison, ResNet-10~\cite{he2016deep} is selected as our feature extractor $\mathcal{F_\theta}$, and GNN~\cite{garcia2017few} is selected as our FSL classifier $\mathcal{G_\phi}$. We divide the Resnet-10 into 4 blocks and one last fully connected (FC) layer, and insert our DWT, $\mathcal{M}_{SA}$, and IDWT after block1, block 2, and block 3.  Our SSL module $\mathcal{M}_{SSL}$ is applied after the FSL classifier.

\noindent\textbf{Implementation Details.} 
We adopt a two-stage training strategy. Firstly, the  $\mathcal{F_\theta}$ is pretrained by performing the standard classification task which is minimized \fyq{by} the cross-entropy loss. Then, we meta-train the whole model. Again, only source data is used for both training stages. We set the $k_1=0.2$, $k_2=0.8$. 
Adam is used as the optimizer. It takes 400 epochs for the pretraining stage and 200 episodes for the meta-training stage.
	
\noindent\textbf{Evaluation Metric.} We conduct experiments on 5-way-1-shot and 5-way-5-shot settings. In testing, 1000 episodes are randomly sampled from the target dataset to evaluate the model. The mean classification results are reported.

\subsection{Main Results}
	
\noindent\textbf{Baselines and Competitors.} 
We compare our wave-SAN with three groups of baselines and competitors: (1) classical few-shot learning (FSL) methods; (2) data augmentation and SSL (Aug $\&$ SSL) methods; (3) cross-domain few-shot learning (CD-FSL) methods. For classical FSL methods, three metric-learning based methods: RelationNet~\cite{sung2018learning}, MatchingNet~\cite{vinyals2016matching}, and GNN~\cite{garcia2017few} are included. 

For Aug $\&$ SSL methods, following the basic idea of data-augmentation based self/semi-supervised learning methods -- augmenting the input data while keeping their semantic labels unaffected, e.g. FixMatch~\cite{sohn2020fixmatch}, we propose several naive but reasonable competitors: Gaussian Noise, ImgAug-weak, and ImgAug-strong. Besides, the MixStyle~\cite{zhou2021domain} is also considered.
The pipeline for these competitors is similar to ours, where the ResNet-10 is used as the feature extractor and GNN is used as the FSL classifier. An episode $A$ is randomly sampled per training step, then augmentation is applied to generate a new episode $A_{aug}$.  Both $A$ and $A_{aug}$ are used to conduct FSL tasks resulting in the predictions $P_{A_0}$, $P_{A_{aug}}$ and losses $\mathcal{L}_{A_{0}}$ and $\mathcal{L}_{A_{aug}}$.  Besides, we use our $\mathcal{M}_{SSL}$ to calculate the KL loss $\mathcal{L}_{A_{SSL}}$ between $P_{A_0}$ and $P_{A_{aug}}$. Finally, $\mathcal{L}_{A_0}$, $\mathcal{L}_{A_{aug}}$, and $\mathcal{L}_{A_{SSL}}$ are used to meta-train the model together. For the Gaussian Noise baseline, we randomly add the Gaussian noises to the input images of the episode $A$. For the ImgAug-weak and ImgAug-strong, a combination of random cropping, color jittering, random rotation, random grayscale, and gaussian blur is applied. The ImgAug-strong has a higher degree of augmentation than ImgAug-weak. As MixStyle~\cite{zhou2021domain} is originally proposed for domain adaption, we adapt it for CD-FSL in the comparison. The implementation details are in \fyq{supplementary materials}.

For CD-FSL methods, four recently proposed methods including FWT~\cite{tseng2020cross}, LRP~\cite{sun2021explanation}, ATA~\cite{wang2021cross}, and Meta-FDMixup~\cite{fu2021meta} are introduced as our competitors. All of them are built on the basis of ResNet-10 and GNN. For fair comparisons, all these methods are implemented under the same setting as our method, i.e., training with single source domain. Note that despite the original Meta-FDMixup using both source and target data for training, it can be directly adapted to our setting by simply sampling two episodes from the source dataset as the input.
In addition, there are also several other CD-FSL methods, e.g., STARTUP~\cite{phoo2020self}, TriAE~\cite{guan2020large}, ConFT~\cite{das2021importance}, and NSAE~\cite{liang2021boosting}. These methods either require information from target domains (e.g., unlabeled target images, semantic labels) for training or require target images for fine-tuning \fyq{during inference}, which makes it unfair to compare our model with them. Hence, we do not make these methods as our competitors.

	\begin{table*}[!t] \footnotesize
		\begin{center}
			\begin{tabular} { c c c c c c}
				\toprule
				\textbf{1-shot} & \textbf{mini} & \textbf{CUB} & \textbf{Cars} & \textbf{Places} & \textbf{Plantae} \\
				\hline       
				
				%&CE Train~\cite{das2021importance} & - &  43.42\mypm0.75 & \textcolor{blue}{35.19\mypm0.66} & 49.56\mypm0.80 & \textcolor{blue}{40.39\mypm0.79} \\
				
				RelationNet~\cite{sung2018learning} & 57.80\mypm0.88 & 42.44\mypm0.77 & 29.11\mypm0.60 & 48.64\mypm0.85 & 33.17\mypm0.64  \\
				
				MatchingNet~\cite{vinyals2016matching} & 59.10\mypm0.64 & 35.89\mypm0.51 & 30.77\mypm0.47 & 49.86\mypm0.79 & 32.70\mypm0.60  \\
				
				GNN~\cite{garcia2017few} &  60.77\mypm0.75 & 45.69\mypm0.68 & 31.79\mypm0.51 & 53.10\mypm0.80 & 35.60\mypm0.56 \\
				\hline

				%Aug $\&$ SSL 
				Gaussian Noise$\dagger$  & 58.59\mypm0.78	& 44.10\mypm0.71	& 30.58\mypm0.54	& 50.72\mypm0.75	& 36.26\mypm0.58 \\
				
				ImgAug-weak$\dagger$     & 64.85\mypm0.80 & 43.96\mypm0.69 & 30.99\mypm0.51 & 56.04\mypm0.79 & 37.18\mypm0.62\\
				
				ImgAug-strong$\dagger$  & 64.87\mypm0.82 & 43.38\mypm0.66 & 29.35\mypm0.47 & 55.83\mypm0.80 & 36.55\mypm0.60\\
				
				%& Mixup~\cite{zhang2017mixup}$\dagger$ & 63.58\mypm0.80	& 46.86\mypm0.74	& 31.61\mypm0.55	& 54.45\mypm0.79	& 36.57\mypm0.58	 \\
				
				%MixStyle~\cite{zhou2021domain}$\dagger$   & 66.08\mypm0.79 & 47.63\mypm0.74 & 32.07\mypm0.57 & 57.13\mypm0.81 & 38.10\mypm0.62 \\
				
				MixStyle~\cite{zhou2021domain}$\dagger$   & 64.86\mypm0.79 & 47.08\mypm0.73 & 33.39\mypm0.58 & 56.12\mypm0.78 & 38.03\mypm0.62 \\

				\hline

				%IN& IN-PFalse~\cite{ulyanov2017improved}$\dagger$  &  42.84\mypm0.65 & 34.45\mypm0.56 & 29.61\mypm0.51 & 40.21\mypm0.65 & 32.90\mypm0.57\\
				
				%&IN-PTrue~\cite{ulyanov2017improved}$\dagger$  & 48.44\mypm0.73 & 39.39\mypm0.65 & 31.74\mypm0.56 & 43.54\mypm0.67 & 34.13\mypm0.58\\
				
				%\hline

				%CD-FSL& 
				FWT~\cite{tseng2020cross}  & 66.32\mypm0.80 & 47.47\mypm0.75 & 31.61\mypm0.53 & 55.77\mypm0.79 & 35.95\mypm0.58 \\
				
				LRP~\cite{sun2021explanation}  & 65.03\mypm0.54 & 48.29\mypm0.51 & 32.78\mypm0.39 & 54.83\mypm0.56  & 37.49\mypm0.43 \\
				
				ATA~\cite{wang2021cross} & - & 45.00\mypm0.50 & \textbf{33.61\mypm0.40} & 53.57\mypm0.50 & 34.42\mypm0.40 \\
				
				Meta-FDMixup~\cite{fu2021meta} $\dagger$ & 62.12\mypm0.76	& 46.38\mypm0.68	& 31.14\mypm0.51	& 53.57\mypm0.75 & 37.89\mypm0.58 \\
				%& CE Train + ConFT~\cite{das2021importance} & - & 45.57\mypm0.76\  & \textcolor{blue}{39.11\mypm0.77} & 49.97\mypm0.86 & \textcolor{blue}{43.09\mypm0.78} \\
				
				\hline
				
				%ours  & 
				\textbf{GNN$+$wave-SAN} & 
				\textbf{67.21\mypm0.79}& \textbf{50.25\mypm0.74} & 
				33.55\mypm0.61 & 
				\textbf{57.75\mypm0.82} & \textbf{40.71\mypm0.66} \\

				\textbf{FWT$+$wave-SAN} & 
				65.95\mypm0.81 & \textbf{50.33\mypm0.73} & 
				32.69\mypm0.59
				& \textbf{57.84\mypm0.81} & \textbf{38.25\mypm0.63} \\

				\hline
				\hline
				
				\textbf{5-shot}  &  \textbf{mini} & \textbf{CUB} & \textbf{Cars} & \textbf{Places} & \textbf{Plantae} \\
				\hline       
				%classical FSL 
				%& CE Train~\cite{das2021importance} & - & 62.80\mypm0.76 & \textcolor{blue}{51.41\mypm0.72}  & 70.71\mypm0.68 &  \textcolor{blue}{55.54\mypm0.69} \\
				
				RelationNet~\cite{sung2018learning} & 71.00\mypm0.69 & 57.77\mypm0.69 & 37.33\mypm0.68 & 63.32\mypm0.76 & 44.00\mypm0.60\\
				
				MatchingNet~\cite{vinyals2016matching} & 70.96\mypm0.65 & 51.37\mypm0.77 & 38.99\mypm0.64 & 63.16\mypm0.77 & 46.53\mypm0.68\\
				
				GNN~\cite{garcia2017few} &  80.87\mypm0.56 & 62.25\mypm0.65 & 44.28\mypm0.63 & 70.84\mypm0.65 & 52.53\mypm0.59\\
				\hline
				
				%Aug $\&$ SSL 
				Gaussian Noise$\dagger$   &  77.19\mypm0.59	& 62.35\mypm0.67	& 41.64\mypm0.61	& 70.09\mypm0.65	& 51.86\mypm0.61 \\
				
				ImgAug-weak$\dagger$      & 82.46\mypm0.55 & 64.53\mypm0.71 & 43.86\mypm0.59 & 75.47\mypm0.66 & 55.39\mypm0.64 \\
				
				ImgAug-strong$\dagger$   & 81.63\mypm0.59 & 61.83\mypm0.71 & 42.15\mypm0.58 & 74.77\mypm0.67 & 53.77\mypm0.64 \\
				
				%& Mixup~\cite{zhang2017mixup}$\dagger$   & 81.37\mypm 0.57	& 65.56\mypm 0.67	& 43.27\mypm0.61&	73.14\mypm0.63	& 53.22\mypm0.59 \\
				
				%MixStyle~\cite{zhou2021domain}$\dagger$   & 83.71\mypm0.54 & 68.53\mypm0.70 &  46.03\mypm0.67 & 76.52\mypm0.64 & 57.39\mypm0.64\\
				
				MixStyle~\cite{zhou2021domain}$\dagger$   & 82.54\mypm0.56 & 65.73 \mypm 0.66 &	45.91\mypm0.63 & 75.90\mypm0.63 & 56.59\mypm0.62  \\
		
				\hline

				%IN & IN-PFalse~\cite{ulyanov2017improved}$\dagger$  & 57.80\mypm0.60 & 47.59\mypm0.61 &  41.27\mypm0.56 & 56.37\mypm0.66 & 44.99\mypm0.59 \\
				
				%& IN-PTrue~\cite{ulyanov2017improved}$\dagger$   & 65.52\mypm0.67 & 53.81\mypm0.66 & 43.63\mypm0.61 & 62.45\mypm0.63 & 48.85\mypm0.62 \\
				
				%\hline

				%CD-FSL 
				FWT~\cite{tseng2020cross}  
				& 81.98\mypm0.55 
				& 66.98\mypm0.68 
				& 44.90\mypm0.64 
				& 73.94\mypm0.67 
				& 53.85\mypm0.62 \\
				
				LRP~\cite{sun2021explanation} 
				& 82.03\mypm0.40 
				& 64.44\mypm0.48 
				& 46.20\mypm0.46 
				& 74.45\mypm0.47 
				& 54.46\mypm0.46 \\
				
				ATA~\cite{wang2021cross} & - & 66.22\mypm0.50 & \textbf{49.14\mypm0.40} & 75.48\mypm0.40 & 52.69\mypm0.40 \\
				
				Meta-FDMixup~\cite{fu2021meta}$\dagger$ & 81.07\mypm 0.55 & 64.71\mypm 0.68 & 41.30\mypm 0.58	& 73.42\mypm0.65	& 54.62\mypm 0.66 \\
				
				%& CE Train + ConFT ~\cite{das2021importance} 
				%& - 
				%& 70.53\mypm0.75 
				%& \textcolor{blue}{61.53\mypm0.75}
				%& 72.09\mypm0.68 
				%& \textcolor{blue}{62.54\mypm0.76} \\
				\hline
				
				%ours & 
				\textbf{GNN$+$wave-SAN} & \textbf{ 84.27\mypm0.54 } & \textbf{ 70.31\mypm0.67 } & 
				46.11\mypm0.66 & 
				\textbf{ 76.88\mypm0.63 } & \textbf{ 57.72\mypm0.64 }\\
						
				\textbf{FWT$+$wave-SAN} & 
				\textbf{ 85.94\mypm0.50 } & \textbf{ 71.16\mypm0.66 } & 
				47.78\mypm0.67 & 
				\textbf{ 78.19\mypm0.62 } & \textbf{ 57.85\mypm0.66} \\		
				\bottomrule
			\end{tabular}
		\end{center}
	    %\vspace{-0.1in}
		\caption{\textbf{Comparative results (\%) for 5-way-1-shot and 5-way-5-shot CD-FSL tasks on FWT's benchmark.} Models are trained on mini-Imagenet (abbreviated to ``mini'') and evaluated on CUB, Cars, Places, and Plantae, respectively.  $\dagger$ indicates the results are reproduced by ourselves. Our method improves the GNN and FWT significantly, and outperforms all the other competitors in most cases.} %outperforms all the other baselines and competitors in most cases.}
		\label{tab-main-result}
		%\vspace{-8mm}
		%\vspace{-0.1in}
	\end{table*}

\noindent\textbf{Results on FWT's Benchmark.} Table~\ref{tab-main-result} summarizes the performance comparison of 5-way-1-shot and 5-way-5-shot CD-FSL tasks on FWT's benchmark. 
From the results, we have the following observations. 1) Firstly, our wave-SAN models outperform all the classical FSL methods, Aug $\&$ SSL methods, and CD-FSL methods in most cases. Specifically, our ``GNN + wave-SAN'' achieves 50.25\%, 33.55\%, 57.75\%, and 40.71\% on CUB, Cars, Places, and Plantae under the 5-way-1-shot setting, improving the GNN baseline up to 4.56\%, 1.76\%, 4.65\%, and 5.11\%, respectively. The superior performance of our methods demonstrates that the proposed style expansion is effective in narrowing the domain gap between the source and target datasets in CD-FSL. 
In addition, our ``GNN + wave-SAN'' also achieves the best results on the source test set (i.e.,mini-Imagenet). Concretely, compared to GNN, our wave-SAN improves the performance by 6.44\% and 3.40\% under 1-shot and 5-shot settings, respectively. The results indicate that our style augmentation does not damage the original semantic information in images while expanding their styles. 
2) Secondly, a performance improvement can be observed by comparing the results of ``FWT + wave-SAN'' with that of FWT in most cases. Applying our key modules, for the 5-way-5-shot CD-FSL, the performances of the FWT have been improved by 4.18\%, 2.88\%, 4.25\%, and 4.00\% on CUB, Cars, Places, and Plantae, respectively. The results demonstrate that our method is complementary to existing CD-FSL methods, e.g., FWT. 
3) Thirdly, among different Aug $\&$ SSL methods, MixStyle which mixes the styles of two episodes achieves the best results. ``Gaussian Noise'', in contrast, performs the worst among all the data augmentation methods. This demonstrates that augmenting the styles is more effective in addressing the CD-FSL problem compared to other global augmentation methods. Another point worth mentioning is that though both our method and MixStyle perform style augmentations, the performance of MixStyle is inferior to our wave-SAN. The results basically indicate the superiority of augmenting styles upon low-frequency components rather than overall features. 

As for the CD-FSL competitors, in general, all the FWT, LRP, ATA, and MetaMixup perform better than the GNN. The performances of FWT, LRP, and ATA are not difficult to understand since they are specifically designed for recognizing novel classes for CD-FSL with only a source dataset. We note that the Meta-FDMixup also improves the GNN in many cases when the input episodes are sampled from source images only. This demonstrates that the source domain itself has a certain degree of visual diversity. This inherent but limited diversity is well explored by our model by taking advantage of them to augment the source styles. When comparing our wave-SAN to these CD-FSL competitors, our method shows obvious advantages in most cases. 
Our superiority on other target datasets in turn shows the most key challenge of CD-FSL lies in the style shift between source and target domains. By tackling this problem directly, our method improves the generalization ability of the model more effectively.

\noindent\textbf{Results on BSCD-FSL Benchmark.} 
We further compare our wave-SAN against FSL and CD-FSL competitors on the BSCD-FSL benchmark. Results are reported in Table~\ref{tab:bscd}. Overall, the target datasets in this benchmark are more challenging than those of FWT's benchmark. Especially, the most challenging target dataset - ChestX can only have the recognition accuracy at 25.63\%. This puts higher demands on the generalization ability of the models.

\begin{table*}[h]\small
%\vspace{-0.1in}
\begin{center}
\begin{tabular}{lcccc}
\toprule
\textbf{1-shot}
& \textbf{ChestX}        
& \textbf{ISIC}
& \textbf{EuroSAT}
& \textbf{CropDisease}  \\ \hline
MatchingNet~\cite{vinyals2016matching}$\dagger$& 20.91\mypm0.30 & 29.46\mypm0.56 & 50.67\mypm0.88 & 48.47\mypm1.01 \\

RelationNet~\cite{sung2018learning}$\dagger$ & 21.94\mypm0.42 & 29.69\mypm0.60 & 56.28\mypm0.82 & 56.18\mypm0.85 \\

GNN~\cite{garcia2017few}$\dagger$ & 22.00\mypm0.46  & 32.02\mypm0.66 & 63.69\mypm1.03 & 64.48\mypm1.08
\\ 
\hline

%MixStyle~\cite{zhou2021domain}$\dagger$ & 22.43\mypm0.35 & 33.21\mypm0.53 & 67.35\mypm0.80 & 68.80\mypm0.82\\ \hline

FWT~\cite{tseng2020cross}$\dagger$ & 22.04\mypm0.44 & 31.58\mypm0.67 & 62.36\mypm1.05 & 66.36\mypm1.04
\\ 

LRP~\cite{sun2021explanation} & 22.11\mypm0.20 &  30.94\mypm0.30 & 54.99\mypm0.50 & 59.23\mypm0.50
\\

ATA~\cite{wang2021cross} & 22.10\mypm0.20 & 33.21\mypm0.40 & 61.35\mypm0.50 & 67.47\mypm0.50 
\\

Meta-FDMixup~\cite{fu2021meta}$\dagger$ & 22.26\mypm 0.45 &	32.48\mypm0.64 &  62.97\mypm 1.01 &	66.23\mypm1.03 
\\
\hline

%\textbf{GNN+StyleAdv} & 22.59\mypm0.35 & 33.64\mypm0.54 & 67.88\mypm0.83 & 70.82\mypm0.79 \\
\textbf{GNN$+$wave-SAN} 
             & \textbf{22.93\mypm0.49} & \textbf{33.35\mypm0.71} & \textbf{69.64\mypm1.09} & \textbf{70.80\mypm1.06}  \\
\textbf{FWT$+$wave-SAN} 
             & \textbf{22.39\mypm0.46} & 
             33.09\mypm0.69
             &  \textbf{65.50\mypm1.09} &  \textbf{69.65\mypm1.04} \\ 

\hline
\hline

\textbf{5-shot} & \textbf{ChestX}  & \textbf{ISIC} & \textbf{EuroSAT} & \textbf{CropDisease} \\ \hline

MatchingNet~\cite{vinyals2016matching}$\dagger$& 22.56\mypm0.36 & 34.38\mypm0.52 &  66.80\mypm0.76 & 58.73\mypm1.03 \\

RelationNet~\cite{sung2018learning}$\dagger$ & 24.34\mypm0.41 & 37.07\mypm0.52 & 68.08\mypm0.69 & 75.33\mypm0.71 \\

GNN~\cite{garcia2017few}$\dagger$ & 25.27\mypm0.46 &  43.94\mypm0.67 & 83.64\mypm0.77 & 87.96\mypm0.67 \\ 
\hline

%MixStyle~\cite{zhou2021domain}$\dagger$ & 25.04\mypm0.36 & 43.77\mypm0.53 & 82.67\mypm0.58 & 88.90\mypm0.52\\ \hline

FWT~\cite{tseng2020cross}$\dagger$ & 25.18\mypm0.45  & 43.17\mypm0.70 & 83.01\mypm0.79 & 87.11\mypm0.67 \\ 

LRP~\cite{sun2021explanation} & 24.53\mypm0.30 & 44.14\mypm0.40 & 77.14\mypm0.40  & 86.15\mypm0.40 \\

ATA~\cite{wang2021cross} & 24.32\mypm0.40 & 44.91\mypm0.40 & 83.75\mypm0.40 & 90.59\mypm0.30 \\

Meta-FDMixup~\cite{fu2021meta}$\dagger$ & 24.52\mypm0.44  & 44.28\mypm 0.66 & 80.48\mypm 0.79 & 87.27\mypm0.69 \\
\hline
%\textbf{GNN+StyleAdv} & 25.12\mypm0.38 & 45.40\mypm0.52 &	86.32\mypm0.55 & 91.55\mypm0.49 \\
\textbf{GNN$+$wave-SAN} 
             & \textbf{25.63\mypm0.49} & \textbf{44.93\mypm0.67} & \textbf{85.22\mypm0.71} &
             89.70\mypm0.64 \\
             
\textbf{FWT$+$wave-SAN} 
             & 
             \textbf{25.27\mypm0.47} 
             & \textbf{46.00\mypm0.72}  & \textbf{84.84\mypm0.68} & \textbf{91.23\mypm0.53} \\ 
\bottomrule
\end{tabular}
\end{center}
%\vspace{-0.1in}
\caption{ \textbf{Comparative results (\%) on  BSCD-FSL benchmark.} 5-way-1-shot and 5-way-5-shot settings are conducted. $\dagger$ indicates the results are reproduced by ourselves. Among all the baselines and competitors, our wave-SAN models based on the GNN and the FWT achieves the best result.}
%\vspace{-0.15in}
\label{tab:bscd}
\end{table*}

Under such a huge domain gap, the performance improvements gained from existing CD-FSL methods (i.e., FWT, LRP, ATA, Meta-FDMixup) 
are somehow limited and even some dropping can be observed. This may be because that they need to take advantage of the visual correlation between source and target more or less. For example, FWT expects that the hyperparameters of batch normalization layers manually tuned on the source domain can still work for the target datasets; LRP relies on the results of the explanation map generated by~\cite{bach2015pixel}, which is highly related to the visual information. 
In contrast, under the 5-way-1-shot setting, our ``GNN + wave-SAN'' still improves the GNN by 0.93\%, 1.33\%, 5.95\%, and 6.32\% for ChestX, ISIC, EuroSAT, and CropDisease datasets, respectively. This indicates that no matter what the target dataset is, the style shift problem is the fundamental issue for CD-FSL. By tackling this problem directly, our ``GNN + wave-SAN'' and ``FWT + wave-SAN'' boost the vanilla GNN and FWT steadily on different target datasets.

%\subsection{Ablation Study}
\subsection{Ablation Studies}
%\fyq{\noindent\textbf{Ablation Study on Each Module.}}
We conduct experiments to show the effectiveness of each proposed component, especially the necessity of introducing wavelet transforms into our method. The results on the FWT's benchmark under the 5-way-1-shot setting are reported in Table~\ref{tab:modules}. The base model is GNN (first line in the table). For the second line without DWT $\&$ IDWT, we apply our $\mathcal{M}_{SA}$ upon the overall feature representation extracted from the blocks. For a fair comparison, it is inserted after the first three blocks. The third line and the last line indicate our model without $\mathcal{M}_{SSL}$ and our full wave-SAN model, respectively.

		\begin{table*}[h]\small
		\begin{center}
		%\vspace{-0.1in}
			\begin{tabular} {c c c c c c c}
				\toprule
				 DWT $\&$ IDWT & \textbf{$\mathcal{M}_{SA}$ }&   \textbf{$\mathcal{M}_{SSL}$} & \textbf{CUB} & \textbf{Cars} & \textbf{Places} & \textbf{Plantae} \\
				\hline
			     \XSolidBrush  &  \XSolidBrush  &  \XSolidBrush  &  45.69\mypm0.68 
			     & 31.79\mypm0.51 & 53.10\mypm0.80 & 35.60\mypm0.56 \\
				\hline
				 
				 \XSolidBrush  &  \Checkmark  &  \Checkmark & 47.99\mypm0.72 & 33.00\mypm0.57 & 56.71\mypm0.83 & 38.44\mypm0.61 \\ 
				\hline
				
				\Checkmark & \Checkmark  & \XSolidBrush & 48.83\mypm0.72 & 33.01\mypm0.59 	& 56.99\mypm0.80 & 38.97\mypm0.64 \\
				\hline
				
				\Checkmark &  \Checkmark & \Checkmark & \textbf{50.25\mypm0.74} & \textbf{33.55\mypm0.61} & \textbf{57.75\mypm0.82} & \textbf{40.71\mypm0.66} \\
				
				\bottomrule
			\end{tabular}
		\end{center}
		%\vspace{-0.1in}
		\caption{\textbf{Effectiveness of each component of our method. } We report the results (\%) on the FWT's benchmark under the 5-way-1-shot setting. }
	%\vspace{-0.2in}
	\label{tab:modules}
	\end{table*}
	
Overall, the full wave-SAN model performs best among all the variations. The contribution of the wavelet transforms can be concluded by comparing the second line with our full model. These two comparative experiments demonstrate that augmenting the styles upon the low-frequency components is better than upon the overall features. This further justifies the advantage of keeping the high-frequency components unchanged. In addition, since the wavelet transforms are completely reversible, applying DWT $\&$ IDWT alone is equivalent to the GNN. Thus, by comparing the first line and the third line, we show the effectiveness of our $\mathcal{M}_{SA}$. The $\mathcal{M}_{SSL}$ also contributes to our full model which can be observed by comparing the last two lines.

Furthermore, we conduct more ablation experiments to evaluate our method. Particularly, in the supplementary materials, we include the experiments of class selection strategy for dual episodes, using different wavelet transforms, trying which blocks are the best to augment the styles, and validating the choices of the hyper-parameters $k_1$ and $k_2$, etc.

\subsection{Visualization}

\begin{figure*}[h]
\centering
\includegraphics[width=0.95\linewidth]{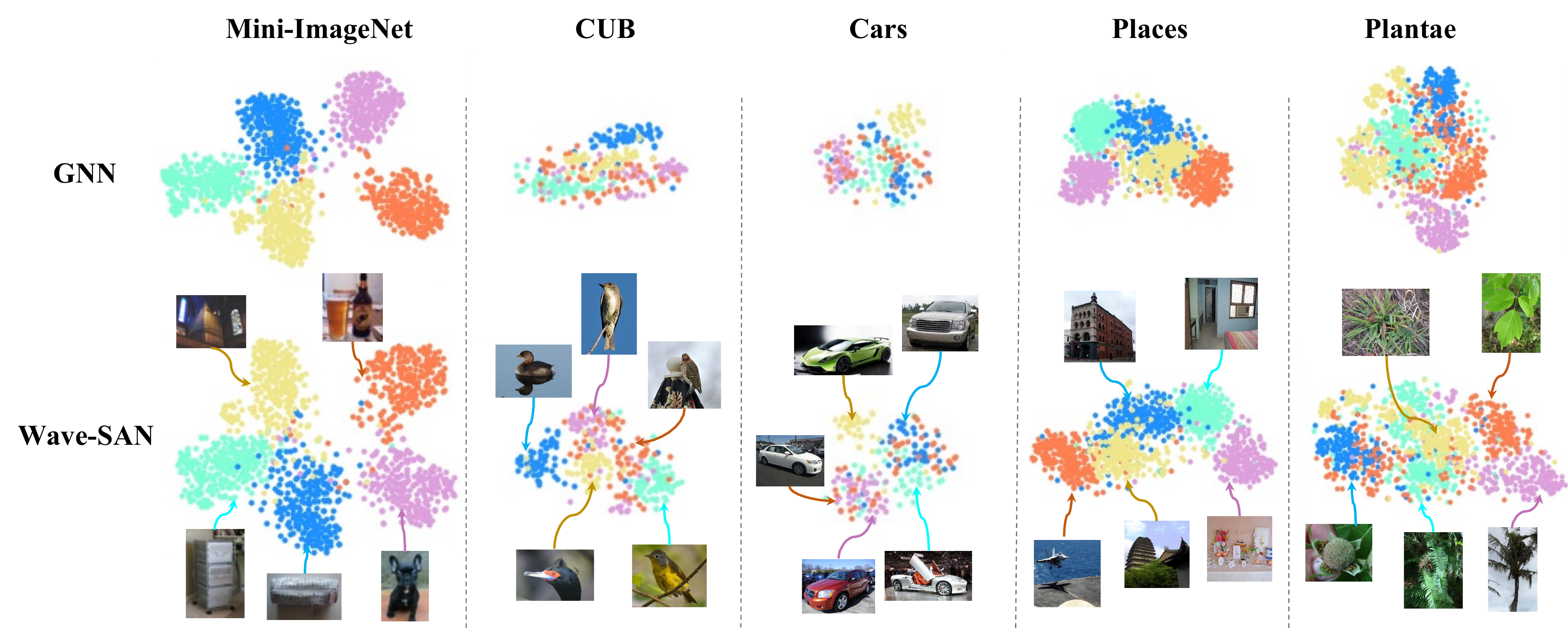}
%\vspace{-0.1in}
\caption{\textbf{T-SNE visualization results.} \small A comparison of our wave-SAN and GNN on the mini-Imagenet and four target datasets is given. For each dataset, five categories are randomly sampled. Different colors indicate different categories.}
\label{fig:t-sne} 
%\vspace{-0.15in}
\end{figure*}

To intuitively show how our method improves the generalization ability of the FSL model, we visualize the learned features via t-SNE in Figure~\ref{fig:t-sne}. Concretely, the representations encoded by the feature extractor are projected into a 2D space. We compare our GNN based wave-SAN against the vanilla GNN model. 

From the results, it can be observed that the overall distributions of these two models are similar, which indicates our method doesn't affect the semantic label of the source images. Besides, compared to GNN, the distribution of each category are indeed more dispersed in wave-SAN as expected. In addition, for target datasets, our wave-SAN enlarges inter-class distances (e.g., CUB, Cars, Places, and Plantae) and reduces the intra-class distances (e.g. CUB), hence making it easier for the model to recognize the target categories.

%%%%%%%%% conclusion
\section{Conclusion}
In this paper, we propose to tackle the data shift problem of CD-FSL by spanning the style distributions of the source domain. This perspective has not been directly investigated in previous works. Technically, a novel wave-SAN model which is mainly composed of wavelet transforms, a StyleAug module, and an SSL module is proposed. These modules together augment the source styles without the requirement of learnable parameters. Experimental results demonstrate that our model improves the generalization ability of FSL models and also boosts other CD-FSL models in a plug-and-play manner. Furthermore, our model has the merit of dealing with large visual styles shift.
Principally, the core idea of our method can be used as a general solution to narrow the domain gap. However, we have not yet conducted experiments to further validate this for general domain adaptation. This could be our future work.

%%%%%%%%%%%%%%%%%%%%%%%% refer
\clearpage
\printbibliography

@article{chen2019closer,
	title={A closer look at few-shot classification},
	author={Chen, Wei-Yu and Liu, Yen-Cheng and Kira, Zsolt and Wang, Yu-Chiang Frank and Huang, Jia-Bin},
	journal = {arXiv preprint},
	year={2019}
}

@inproceedings{snell2017prototypical,
	title={Prototypical networks for few-shot learning},
	author={Snell, Jake and Swersky, Kevin and Zemel, Richard},
	booktitle = {NeurIPS},
	year={2017}
}

@inproceedings{vinyals2016matching,
	title={Matching networks for one shot learning},
	author={Vinyals, Oriol and Blundell, Charles and Lillicrap, Timothy and Wierstra, Daan and others},
	booktitle = {NeurIPS},
	year={2016}
}

@inproceedings{sung2018learning,
	title={Learning to compare: Relation network for few-shot learning},
	author={Sung, Flood and Yang, Yongxin and Zhang, Li and Xiang, Tao and Torr, Philip HS and Hospedales, Timothy M},
	booktitle = {CVPR},
	year={2018}
}

@article{garcia2017few,
	title={Few-shot learning with graph neural networks},
	author={Garcia, Victor and Bruna, Joan},
	journal = {arXiv preprint},
	year={2017}
}

@inproceedings{finn2017model,
  title={Model-agnostic meta-learning for fast adaptation of deep networks},
  author={Finn, Chelsea and Abbeel, Pieter and Levine, Sergey},
  booktitle={International Conference on Machine Learning},
  pages={1126--1135},
  year={2017},
  organization={PMLR}
}

@inproceedings{munkhdalai2017meta,
  title={Meta networks},
  author={Munkhdalai, Tsendsuren and Yu, Hong},
  booktitle={International Conference on Machine Learning},
  pages={2554--2563},
  year={2017},
  organization={PMLR}
}

@article{rusu2018meta,
  title={Meta-learning with latent embedding optimization},
  author={Rusu, Andrei A and Rao, Dushyant and Sygnowski, Jakub and Vinyals, Oriol and Pascanu, Razvan and Osindero, Simon and Hadsell, Raia},
  journal={arXiv preprint arXiv:1807.05960},
  year={2018}
}

@inproceedings{chen2019image,
	title={Image block augmentation for one-shot learning},
	author={Chen, Zitian and Fu, Yanwei and Chen, Kaiyu and Jiang, Yu-Gang},
	booktitle={AAAI},
	year={2019}
}

@inproceedings{li2020adversarial,
	title={Adversarial Feature Hallucination Networks for Few-Shot Learning},
	author={Li, Kai and Zhang, Yulun and Li, Kunpeng and Fu, Yun},
	booktitle = {CVPR},
	year={2020}
}

@inproceedings{hariharan2017low,
  title={Low-shot visual recognition by shrinking and hallucinating features},
  author={Hariharan, Bharath and Girshick, Ross},
  booktitle={Proceedings of the IEEE International Conference on Computer Vision},
  pages={3018--3027},
  year={2017}
}

@inproceedings{tseng2020cross,
	title={Cross-domain few-shot classification via learned feature-wise transformation},
	author={Tseng, Hung-Yu and Lee, Hsin-Ying and Huang, Jia-Bin and Yang, Ming-Hsuan},
	booktitle = {ICLR},
	year={2020}
}

@article{phoo2020self,
	title={Self-training for Few-shot Transfer Across Extreme Task Differences},
	author={Phoo, Cheng Perng and Hariharan, Bharath},
	journal = {arXiv preprint},
	year={2020}
}

@inproceedings{fu2021meta,
  title={Meta-FDMixup: Cross-Domain Few-Shot Learning Guided by Labeled Target Data},
  author={Fu, Yuqian and Fu, Yanwei and Jiang, Yu-Gang},
  booktitle={Proceedings of the 29th ACM International Conference on Multimedia},
  pages={5326--5334},
  year={2021}
}

@inproceedings{guo2020broader,
	title={A broader study of cross-domain few-shot learning},
	author={Guo, Yunhui and Codella, Noel C and Karlinsky, Leonid and Codella, James V and Smith, John R and Saenko, Kate and Rosing, Tajana and Feris, Rogerio},
	booktitle = {ECCV},
	year={2020}
}

@inproceedings{sun2021explanation,
  title={Explanation-guided training for cross-domain few-shot classification},
  author={Sun, Jiamei and Lapuschkin, Sebastian and Samek, Wojciech and Zhao, Yunqing and Cheung, Ngai-Man and Binder, Alexander},
  booktitle={2020 25th International Conference on Pattern Recognition (ICPR)},
  pages={7609--7616},
  year={2021},
  organization={IEEE}
}

@inproceedings{liang2021boosting,
  title={Boosting the Generalization Capability in Cross-Domain Few-shot Learning via Noise-enhanced Supervised Autoencoder},
  author={Liang, Hanwen and Zhang, Qiong and Dai, Peng and Lu, Juwei},
  booktitle={Proceedings of the IEEE/CVF International Conference on Computer Vision},
  pages={9424--9434},
  year={2021}
}

@article{yeh2020large,
  title={Large margin mechanism and pseudo query set on cross-domain few-shot learning},
  author={Yeh, Jia-Fong and Lee, Hsin-Ying and Tsai, Bing-Chen and Chen, Yi-Rong and Huang, Ping-Chia and Hsu, Winston H},
  journal={arXiv preprint arXiv:2005.09218},
  year={2020}
}

@article{liu2020feature,
  title={Feature transformation ensemble model with batch spectral regularization for cross-domain few-shot classification},
  author={Liu, Bingyu and Zhao, Zhen and Li, Zhenpeng and Jiang, Jianan and Guo, Yuhong and Ye, Jieping},
  journal={arXiv preprint arXiv:2005.08463},
  year={2020}
}

@inproceedings{das2021importance,
  title={On the Importance of Distractors for Few-Shot Classification},
  author={Das, Rajshekhar and Wang, Yu-Xiong and Moura, Jose MF},
  booktitle={Proceedings of the IEEE/CVF International Conference on Computer Vision},
  pages={9030--9040},
  year={2021}
}

@inproceedings{lee2019meta,
  title={Meta-learning with differentiable convex optimization},
  author={Lee, Kwonjoon and Maji, Subhransu and Ravichandran, Avinash and Soatto, Stefano},
  booktitle={Proceedings of the IEEE/CVF Conference on Computer Vision and Pattern Recognition},
  pages={10657--10665},
  year={2019}
}

@article{wang2021cross,
  title={Cross-Domain Few-Shot Classification via Adversarial Task Augmentation},
  author={Wang, Haoqing and Deng, Zhi-Hong},
  journal={arXiv preprint arXiv:2104.14385},
  year={2021}
}

@inproceedings{chen2020simple,
	title={A simple framework for contrastive learning of visual representations},
	author={Chen, Ting and Kornblith, Simon and Norouzi, Mohammad and Hinton, Geoffrey},
	booktitle={International conference on machine learning},
	pages={1597--1607},
	year={2020},
	organization={PMLR}
}

@article{bach2015pixel,
	title={On pixel-wise explanations for non-linear classifier decisions by layer-wise relevance propagation},
	author={Bach, Sebastian and Binder, Alexander and Montavon, Gr{\'e}goire and Klauschen, Frederick and M{\"u}ller, Klaus-Robert and Samek, Wojciech},
	journal={PloS one},
	volume={10},
	number={7},
	pages={e0130140},
	year={2015},
	publisher={Public Library of Science}
}

@inproceedings{guan2020large,
	title={Large-Scale Cross-Domain Few-Shot Learning},
	author={Guan, Jiechao and Zhang, Manli and Lu, Zhiwu},
	booktitle={Proceedings of the Asian Conference on Computer Vision},
	year={2020}
}

@article{adler2020cross,
	title={Cross-Domain Few-Shot Learning by Representation Fusion},
	author={Adler, Thomas and Brandstetter, Johannes and Widrich, Michael and Mayr, Andreas and Kreil, David and Kopp, Michael and Klambauer, G{\"u}nter and Hochreiter, Sepp},
	journal={arXiv preprint arXiv:2010.06498},
	year={2020}
}

@article{islam2021dynamic,
  title={Dynamic distillation network for cross-domain few-shot recognition with unlabeled data},
  author={Islam, Ashraful and Chen, Chun-Fu Richard and Panda, Rameswar and Karlinsky, Leonid and Feris, Rogerio and Radke, Richard},
  journal={Advances in Neural Information Processing Systems},
  volume={34},
  year={2021}
}

@inproceedings{cai2021damsl,
  title={DAMSL: Domain Agnostic Meta Score-based Learning},
  author={Cai, John and Cai, Bill and Mei, Shen Sheng},
  booktitle={Proceedings of the IEEE/CVF Conference on Computer Vision and Pattern Recognition},
  pages={2591--2595},
  year={2021}
}

@inproceedings{zhang2021shallow,
  title={Shallow bayesian meta learning for real-world few-shot recognition},
  author={Zhang, Xueting and Meng, Debin and Gouk, Henry and Hospedales, Timothy M},
  booktitle={Proceedings of the IEEE/CVF International Conference on Computer Vision},
  pages={651--660},
  year={2021}
}

@inproceedings{chen2021self,
  title={Self-supervised learning for few-shot image classification},
  author={Chen, Da and Chen, Yuefeng and Li, Yuhong and Mao, Feng and He, Yuan and Xue, Hui},
  booktitle={ICASSP 2021-2021 IEEE International Conference on Acoustics, Speech and Signal Processing (ICASSP)},
  pages={1745--1749},
  year={2021},
  organization={IEEE}
}

@inproceedings{huang2017arbitrary,
	title={Arbitrary style transfer in real-time with adaptive instance normalization},
	author={Huang, Xun and Belongie, Serge},
	booktitle={Proceedings of the IEEE International Conference on Computer Vision},
	pages={1501--1510},
	year={2017}
}

@inproceedings{zhou2021domain,
	title={Domain generalization with mixstyle},
	author={Zhou, Kaiyang and Yang, Yongxin and Qiao, Yu and Xiang, Tao},
	booktitle={International Conference on Learning Representations},
	year={2021}
}

@article{krizhevsky2012imagenet,
	title={Imagenet classification with deep convolutional neural networks},
	author={Krizhevsky, Alex and Sutskever, Ilya and Hinton, Geoffrey E},
	journal={Advances in neural information processing systems},
	volume={25},
	pages={1097--1105},
	year={2012}
}

@inproceedings{zeiler2014visualizing,
	title={Visualizing and understanding convolutional networks},
	author={Zeiler, Matthew D and Fergus, Rob},
	booktitle={European conference on computer vision},
	pages={818--833},
	year={2014},
	organization={Springer}
}

@article{zhang2017mixup,
	title={mixup: Beyond empirical risk minimization},
	author={Zhang, Hongyi and Cisse, Moustapha and Dauphin, Yann N and Lopez-Paz, David},
	journal = {arXiv preprint},
	year={2017}
}

@inproceedings{verma2019manifold,
	title={Manifold mixup: Better representations by interpolating hidden states},
	author={Verma, Vikas and Lamb, Alex and Beckham, Christopher and Najafi, Amir and Mitliagkas, Ioannis and Lopez-Paz, David and Bengio, Yoshua},
	booktitle = {ICML},
	year={2019}
}

@inproceedings{yun2019cutmix,
	title={Cutmix: Regularization strategy to train strong classifiers with localizable features},
	author={Yun, Sangdoo and Han, Dongyoon and Oh, Seong Joon and Chun, Sanghyuk and Choe, Junsuk and Yoo, Youngjoon},
	booktitle = {ICCV},
	year={2019}
}

@inproceedings{deng2009imagenet,
	title={Imagenet: A large-scale hierarchical image database},
	author={Deng, Jia and Dong, Wei and Socher, Richard and Li, Li-Jia and Li, Kai and Fei-Fei, Li},
	booktitle = {CVPR},
	year={2009}
}

@inproceedings{ravi2016optimization,
	title={Optimization as a model for few-shot learning},
	author={Ravi, Sachin and Larochelle, Hugo},
	booktitle = {ICLR},
	year={2017}
}

@article{wah2011caltech,
	title={The caltech-ucsd birds-200-2011 dataset},
	author={Wah, Catherine and Branson, Steve and Welinder, Peter and Perona, Pietro and Belongie, Serge},
	year={2011}
}

@inproceedings{krause20133d,
	title={3d object representations for fine-grained categorization},
	author={Krause, Jonathan and Stark, Michael and Deng, Jia and Fei-Fei, Li},
	booktitle = {ICCVW},
	year={2013}
}

@article{zhou2017places,
	title={Places: A 10 million image database for scene recognition},
	author={Zhou, Bolei and Lapedriza, Agata and Khosla, Aditya and Oliva, Aude and Torralba, Antonio},
	journal = {TPAMI},
	year={2017}
}

@inproceedings{van2018inaturalist,
	title={The inaturalist species classification and detection dataset},
	author={Van Horn, Grant and Mac Aodha, Oisin and Song, Yang and Cui, Yin and Sun, Chen and Shepard, Alex and Adam, Hartwig and Perona, Pietro and Belongie, Serge},
	booktitle = {CVPR},
	year={2018}
}

@article{mohanty2016using,
  title={Using deep learning for image-based plant disease detection},
  author={Mohanty, Sharada P and Hughes, David P and Salath{\'e}, Marcel},
  journal={Frontiers in plant science},
  volume={7},
  pages={1419},
  year={2016},
  publisher={frontiers}
}

@article{helber2019eurosat,
  title={Eurosat: A novel dataset and deep learning benchmark for land use and land cover classification},
  author={Helber, Patrick and Bischke, Benjamin and Dengel, Andreas and Borth, Damian},
  journal={IEEE Journal of Selected Topics in Applied Earth Observations and Remote Sensing},
  volume={12},
  number={7},
  pages={2217--2226},
  year={2019},
  publisher={IEEE}
}

@article{tschandl2018ham10000,
  title={The HAM10000 dataset, a large collection of multi-source dermatoscopic images of common pigmented skin lesions},
  author={Tschandl, Philipp and Rosendahl, Cliff and Kittler, Harald},
  journal={Scientific data},
  volume={5},
  number={1},
  pages={1--9},
  year={2018},
  publisher={Nature Publishing Group}
}

@article{codella2019skin,
  title={Skin lesion analysis toward melanoma detection 2018: A challenge hosted by the international skin imaging collaboration (isic)},
  author={Codella, Noel and Rotemberg, Veronica and Tschandl, Philipp and Celebi, M Emre and Dusza, Stephen and Gutman, David and Helba, Brian and Kalloo, Aadi and Liopyris, Konstantinos and Marchetti, Michael and others},
  journal={arXiv preprint arXiv:1902.03368},
  year={2019}
}

@inproceedings{wang2017chestx,
  title={Chestx-ray8: Hospital-scale chest x-ray database and benchmarks on weakly-supervised classification and localization of common thorax diseases},
  author={Wang, Xiaosong and Peng, Yifan and Lu, Le and Lu, Zhiyong and Bagheri, Mohammadhadi and Summers, Ronald M},
  booktitle={Proceedings of the IEEE conference on computer vision and pattern recognition},
  pages={2097--2106},
  year={2017}
}

@inproceedings{he2016deep,
	title={Deep residual learning for image recognition},
	author={He, Kaiming and Zhang, Xiangyu and Ren, Shaoqing and Sun, Jian},
	booktitle = {CVPR},
	year={2016}
}

@inproceedings{liu2018multi,
  title={Multi-level wavelet-CNN for image restoration},
  author={Liu, Pengju and Zhang, Hongzhi and Zhang, Kai and Lin, Liang and Zuo, Wangmeng},
  booktitle={Proceedings of the IEEE conference on computer vision and pattern recognition workshops},
  pages={773--782},
  year={2018}
}

@inproceedings{guo2017deep,
  title={Deep wavelet prediction for image super-resolution},
  author={Guo, Tiantong and Seyed Mousavi, Hojjat and Huu Vu, Tiep and Monga, Vishal},
  booktitle={Proceedings of the IEEE Conference on Computer Vision and Pattern Recognition Workshops},
  pages={104--113},
  year={2017}
}

@article{wu2012eulerian,
  title={Eulerian video magnification for revealing subtle changes in the world},
  author={Wu, Hao-Yu and Rubinstein, Michael and Shih, Eugene and Guttag, John and Durand, Fr{\'e}do and Freeman, William},
  journal={ACM transactions on graphics (TOG)},
  volume={31},
  number={4},
  pages={1--8},
  year={2012},
  publisher={ACM New York, NY, USA}
}

@article{zhang1992wavelet,
  title={Wavelet networks},
  author={Zhang, Qinghua and Benveniste, Albert},
  journal={IEEE transactions on Neural Networks},
  volume={3},
  number={6},
  pages={889--898},
  year={1992},
  publisher={IEEE}
}

@article{sohn2020fixmatch,
  title={Fixmatch: Simplifying semi-supervised learning with consistency and confidence},
  author={Sohn, Kihyuk and Berthelot, David and Li, Chun-Liang and Zhang, Zizhao and Carlini, Nicholas and Cubuk, Ekin D and Kurakin, Alex and Zhang, Han and Raffel, Colin},
  journal={arXiv preprint arXiv:2001.07685},
  year={2020}
}

@inproceedings{yoo2019photorealistic,
  title={Photorealistic style transfer via wavelet transforms},
  author={Yoo, Jaejun and Uh, Youngjung and Chun, Sanghyuk and Kang, Byeongkyu and Ha, Jung-Woo},
  booktitle={Proceedings of the IEEE/CVF International Conference on Computer Vision},
  pages={9036--9045},
  year={2019}
}

@inproceedings{yang2020fda,
  title={Fda: Fourier domain adaptation for semantic segmentation},
  author={Yang, Yanchao and Soatto, Stefano},
  booktitle={Proceedings of the IEEE/CVF Conference on Computer Vision and Pattern Recognition},
  pages={4085--4095},
  year={2020}
}

@inproceedings{zou2021sdwnet,
  title={SDWNet: A Straight Dilated Network with Wavelet Transformation for image Deblurring},
  author={Zou, Wenbin and Jiang, Mingchao and Zhang, Yunchen and Chen, Liang and Lu, Zhiyong and Wu, Yi},
  booktitle={Proceedings of the IEEE/CVF International Conference on Computer Vision},
  pages={1895--1904},
  year={2021}
}

@article{min2018blind,
  title={Blind deblurring via a novel recursive deep CNN improved by wavelet transform},
  author={Min, Chao and Wen, Guoquan and Li, Binrui and Fan, Feifei},
  journal={IEEE Access},
  volume={6},
  pages={69242--69252},
  year={2018},
  publisher={IEEE}
}

@inproceedings{gao2019dynamic,
  title={Dynamic scene deblurring with parameter selective sharing and nested skip connections},
  author={Gao, Hongyun and Tao, Xin and Shen, Xiaoyong and Jia, Jiaya},
  booktitle={Proceedings of the IEEE/CVF Conference on Computer Vision and Pattern Recognition},
  pages={3848--3856},
  year={2019}
}

@article{kang2018deep,
  title={Deep convolutional framelet denosing for low-dose CT via wavelet residual network},
  author={Kang, Eunhee and Chang, Won and Yoo, Jaejun and Ye, Jong Chul},
  journal={IEEE transactions on medical imaging},
  volume={37},
  number={6},
  pages={1358--1369},
  year={2018},
  publisher={IEEE}
}

@article{levinskis2013convolutional,
  title={Convolutional neural network feature reduction using wavelet transform},
  author={Levinskis, A},
  journal={Elektronika ir Elektrotechnika},
  volume={19},
  number={3},
  pages={61--64},
  year={2013}
}

@article{gueguen2018faster,
  title={Faster neural networks straight from jpeg},
  author={Gueguen, Lionel and Sergeev, Alex and Kadlec, Ben and Liu, Rosanne and Yosinski, Jason},
  journal={Advances in Neural Information Processing Systems},
  volume={31},
  year={2018}
}

@inproceedings{bae2017beyond,
  title={Beyond deep residual learning for image restoration: Persistent homology-guided manifold simplification},
  author={Bae, Woong and Yoo, Jaejun and Chul Ye, Jong},
  booktitle={Proceedings of the IEEE conference on computer vision and pattern recognition workshops},
  pages={145--153},
  year={2017}
}

@article{fujieda2017wavelet,
  title={Wavelet convolutional neural networks for texture classification},
  author={Fujieda, Shin and Takayama, Kohei and Hachisuka, Toshiya},
  journal={arXiv preprint arXiv:1707.07394},
  year={2017}
}

@inproceedings{ryu2018dft,
  title={Dft-based transformation invariant pooling layer for visual classification},
  author={Ryu, Jongbin and Yang, Ming-Hsuan and Lim, Jongwoo},
  booktitle={Proceedings of the European Conference on Computer Vision (ECCV)},
  pages={84--99},
  year={2018}
}

@inproceedings{williams2018wavelet,
  title={Wavelet pooling for convolutional neural networks},
  author={Williams, Travis and Li, Robert},
  booktitle={International Conference on Learning Representations},
  year={2018}
}

@book{bracewell1986fourier,
  title={The Fourier transform and its applications},
  author={Bracewell, Ronald Newbold and Bracewell, Ronald N},
  volume={31999},
  year={1986},
  publisher={McGraw-hill New York}
}

@inproceedings{deng2019wavelet,
  title={Wavelet domain style transfer for an effective perception-distortion tradeoff in single image super-resolution},
  author={Deng, Xin and Yang, Ren and Xu, Mai and Dragotti, Pier Luigi},
  booktitle={Proceedings of the IEEE/CVF International Conference on Computer Vision},
  pages={3076--3085},
  year={2019}
}

@inproceedings{liu2020remove,
  title={Remove appearance shift for ultrasound image segmentation via fast and universal style transfer},
  author={Liu, Zhendong and Yang, Xin and Gao, Rui and Liu, Shengfeng and Dou, Haoran and He, Shuangchi and Huang, Yuhao and Huang, Yankai and Luo, Huanjia and Zhang, Yuanji and others},
  booktitle={2020 IEEE 17th International Symposium on Biomedical Imaging (ISBI)},
  pages={1824--1828},
  year={2020},
  organization={IEEE}
}

@inproceedings{jamadandi2019exemplar,
  title={Exemplar-based underwater image enhancement augmented by wavelet corrected transforms},
  author={Jamadandi, Adarsh and Mudenagudi, Uma},
  booktitle={Proceedings of the IEEE/CVF Conference on Computer Vision and Pattern Recognition Workshops},
  pages={11--17},
  year={2019}
}

@article{ding2022deep,
  title={Deep attentive style transfer for images with wavelet decomposition},
  author={Ding, Hong and Fu, Gang and Yan, Qinan and Jiang, Caoqing and Cao, Tuo and Li, Wenjie and Hu, Shenghong and Xiao, Chunxia},
  journal={Information Sciences},
  volume={587},
  pages={63--81},
  year={2022},
  publisher={Elsevier}
}

@inproceedings{singh2021safin,
  title={SAFIN: Arbitrary style transfer with self-attentive factorized instance normalization},
  author={Singh, Aaditya and Hingane, Shreeshail and Gong, Xinyu and Wang, Zhangyang},
  booktitle={2021 IEEE International Conference on Multimedia and Expo (ICME)},
  pages={1--6},
  year={2021},
  organization={IEEE}
}

@article{geirhos2018imagenet,
  title={ImageNet-trained CNNs are biased towards texture; increasing shape bias improves accuracy and robustness},
  author={Geirhos, Robert and Rubisch, Patricia and Michaelis, Claudio and Bethge, Matthias and Wichmann, Felix A and Brendel, Wieland},
  journal={arXiv preprint arXiv:1811.12231},
  year={2018}
}

%%%%%%%%%%%%%%%%%%%%%%%%%%%%%%%%%%%%%%%%%%%%%%%%%%%%%%%%%%%%
\clearpage
\appendix

\section{Implementation Details}
%\noindent\textbf{Haar Wavelet Transforms:} 
\subsection{Haar Wavelet Transforms}
Given the input $F$, the 2D Haar wavelet defines four kernels $k_{LL}$, $k_{LH}$, $k_{HL}$, and $k_{HH}$ to extract the low-frequency and high-frequency components.
More specifically, we have:
\begin{gather}
   k_{LL} = \begin{bmatrix} 1 & 1 \\ 1 & 1 \end{bmatrix}, k_{LH} = \begin{bmatrix} -1& -1 \\ 1 & 1 \end{bmatrix},
   k_{HL} = \begin{bmatrix} -1& 1 \\ -1 & 1 \end{bmatrix},
   k_{HH} = \begin{bmatrix} 1& -1 \\ -1 & 1 \end{bmatrix},
\end{gather}
Sliding the kernels across the input $F$ results in four subbands $F_{LL}$, $F_{LH}$, $F_{HL}$, and $F_{HH}$, respectively. In this paper, we have $F_{low} = F_{LL}$, $F_{high} = [F_{LH}, F_{HL}, F_{HH}]$. Due to the reversibility of Haar, the input $F$ can completely be reconstructed by the IDWT. That is, $F = IDWT(F_{low}, F_{high})$.

%\noindent\textbf{Aug $\&$ SSL Competitors.}
\subsection{Aug $\&$ SSL Competitors.}
\begin{figure}[h!]
	\centering
	%\vspace{-0.1in}
	\includegraphics[width=0.85\linewidth]{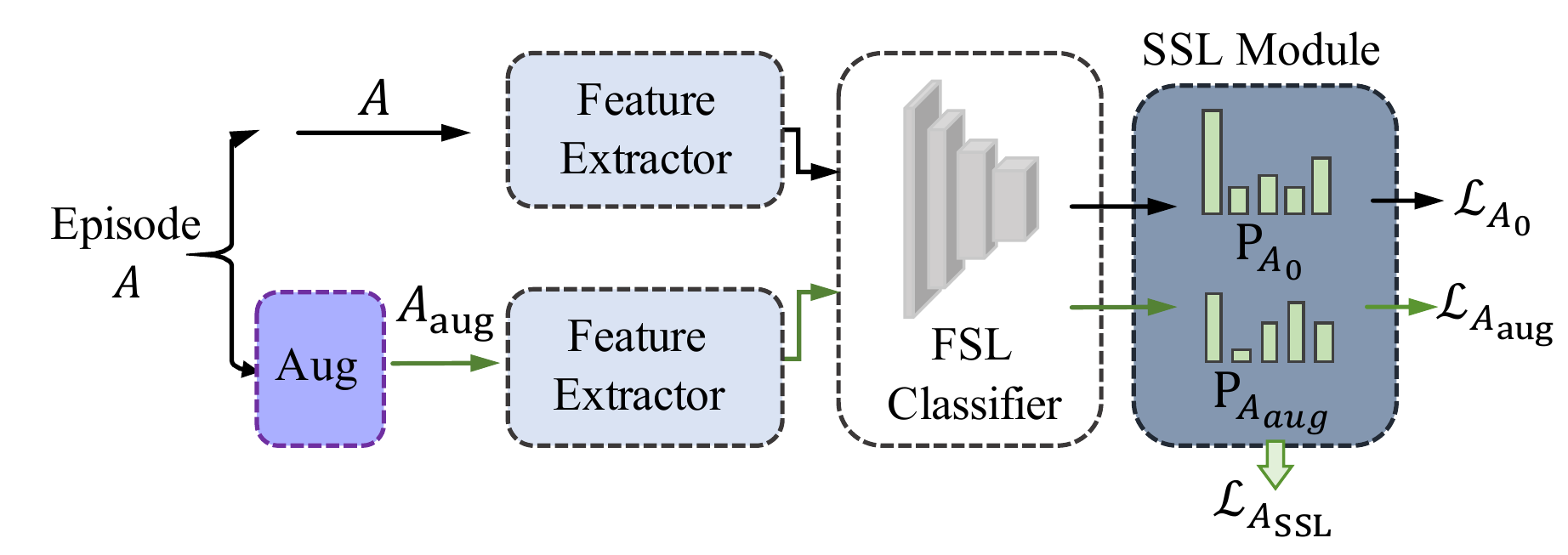}
	%\vspace{-0.1in}
	\caption{\textbf{The pipeline of Aug $\&$ SSL methods.} }
	\label{fig:pipeline} 
	%\vspace{-0.1in}
\end{figure}

We provide the implementation details for the Aug $\&$ SSL competitors compared in our work including ``Gaussian Noise'', ``ImgAug-weak'', ``ImgAug-strong'', and ``MixStyle''. The pipeline for the first three methods is illustrated in Figure~\ref{fig:pipeline}. Overall, for each iteration, we randomly sample an episode $A$ as the input, then specific augmentation is applied to the $A$ resulting in the augmented episode $A_{aug}$. Both the $A$ and the $A_{aug}$ are fed into the feature extractor and the FSL classifier to obtain their predictions $P_{A_{0}}$ and $P_{A_{aug}}$, respectively. Finally, our SSL module is used to calculate the consistency loss between these two predictions generating the $\mathcal{L}_{A_{SSL}}$. These three losses $\mathcal{L}_{A_{0}}$, $\mathcal{L}_{A_{aug}}$, and $\mathcal{L}_{A_{SSL}}$ together help optimize the model. Again, for fair comparisons, ResNet-10 and GNN are adopted as the feature extractor and the FSL classifier, respectively. The details of the augmentation operation for these three methods are given in Table~\ref{aug}.

\begin{table}[h] 
\begin{center}
\begin{tabular}{ll}
\toprule
\textbf{Method} & \textbf{Operations}  \\ \hline
Gaussian Noise               
& 0.2 * Gaussain Noise(1,1,1,224,224)  \\ \hline

ImgAug-weak                
& RandomResizedCrop()                 \\
\multirow{4}{*}{}            
& ImageJitter(s=0.5)                  \\
& RandomRotation(degree=30)           \\
& RandomGrayscale(probability=0.1)    \\
& GaussianBlur()    \\ 
\hline

ImgAug-strong                
& RandomResizedCrop()                 \\
\multirow{4}{*}{}            
& ImageJitter(s=1)                  \\
& RandomRotation(degree=90)           \\
& RandomGrayscale(probability=0.3)    \\
& GaussianBlur()    \\ 
\bottomrule           
\end{tabular}
\end{center}
\caption{\textbf{Augmentation details of our Aug $\&$ SSL methods.}}
\label{aug}
%\vspace{-0.3in}
\end{table}

As for the MixStyle~\cite{zhou2021domain}, since it mixes the styles of two input images, we randomly sample two episodes $A$ and $B$ as the input. Thus, the overall pipeline for MixStyle is similar to ours. We illustrate how the augmentation is done by MixStyle. Formally, take episode $A$ as an example, the style-mixed episode $\Tilde{A}$ is obtained as follows,
	\begin{equation}
	\sigma_{mix} = \lambda \sigma(x_A) + (1-\lambda) \sigma(x_B)
	\end{equation}
	\begin{equation}
	\mu_{mix} = \lambda \mu(x_A) + (1-\lambda) \mu(x_B)
	\end{equation}
	\begin{equation}
	\Tilde{x}_{A} = \sigma_{mix}\frac{x_{A}-\mu(x_{A})}{\sigma(x_{A})} + \mu_{mix}
	\end{equation}
	where the $\sigma$ and $\mu$ means the deviation and the mean of the feature map; the $x_A$ and $x_B$ denotes the features of episode $A$ and $B$, respectively; $\lambda$ is the mixing ratio sampled from the Beta distribution i.e. $\lambda \in Beta(\alpha, \alpha)$. 
	According to its paper, the styles of these two episodes are mixed with a activation probability of $p$, $p=0.5$. Besides, such a mixing operation is applied after the block1, block2, and block3.

\section{Ablation Studies}
\subsection{Analysis of Different Frequency Components.} 
Since the style is mainly conveyed in the low-frequency components, a simple solution is using high-frequency components alone to address the cross domain few-shot classification problem. To demonstrate such simple solution is inferior to the proposed method, we compare our wave-SAN against with the baselines that use high-frequency (HF) components and low-frequency (LF) components alone, respectively. We report the 5-way-5-shot results on FWT's benchmark in Table 2.

\begin{table}[h] \small
%\vspace{-0.1in}
\begin{center}
\begin{tabular} { c c c c c}
	\hline
	\textbf{Methods} & \textbf{Cub} & \textbf{Cars} & \textbf{Places} & \textbf{Plantae} \\
	\hline
	HF only    & 66.38 \mypm 0.68 & 45.42 \mypm 0.64
& 73.74 \mypm 0.65 & 52.44 \mypm 0.59 \\
	\hline
	LF only & 65.95 \mypm 0.68 & 43.36 \mypm 0.62 &  72.15 \mypm 0.66 & 53.21 \mypm 0.60 \\
	\hline
	\textbf{wave-SAN}  & 70.31 \mypm 0.67 & 46.11 \mypm 0.66 & 76.88 \mypm 0.63 & 57.72 \mypm 0.64 \\
	\hline
	\end{tabular}
	\end{center}
	%\vspace{-0.1in}
	\caption{\textbf{Analysis of different frequency components.} 5-way 5-shot results on the FWT's benchmark.}
\label{tab:component}
%\vspace{-0.15in}
\end{table}

The results show that both HF and LF are important for cross domain few-shot recognition, as both HF only and LF only achieve quite high performance. By using both HF and LF well, the proposed wave-SAN outperforms these two baselines that use HF only and LF only. The results suggest despite the ``shift of styles'' in LF leads to domain shift problem, the shape information conveyed in LF is essential for classification. Hence using HF components alone performs worse than our wave-SAN. 

%\fyq{Results show that 1) HF and LF are both important. In other words, it is the \textbf{``shift of styles''} that leads to domain shift problem, while the texture of HF and the main visual information of LF especially the shape are both key to recognizing the categories; 2) The superiority of wave-SAN shows that our method utilizes the HF and LF well.}

\subsection{Performance of different DWT Algorithms.}
As mentioned, in the proposed framework, any wavelet transforms could be used. To demonstrate the proposed wave-SAN works for different wavelet transform algorithms, we further report the performances of our method with several different wavelet transforms, including the ``Daubechies'', ``Coiflets'', ``Biorthogonal'', and ``Symlets''. The 5-way-5-shot results on FWT's benchmark are shown in Table~\ref{tab:waveletAlor}.

%Though we select the most classical Haar algorithm as our wavelet transform method, we also study could our wave-SAN still works for different wavelet transform algorithms? To that end, we perform experiments using several different wavelet algorithms including the ``Daubechies'', ``Coiflets'', ``Biorthogonal'', and ``Symlets''. The 5-way-5-shot results on FWT's benchmark are given in Table~\ref{tab:waveletAlor}.

\begin{table}[h] \small
\begin{center}
%\vspace{-0.1in}
\begin{tabular} { c c c c c}
	\hline
	\textbf{Methods} & \textbf{Cub} & \textbf{Cars} & \textbf{Places} & \textbf{Plantae} \\
	\hline
	GNN~\cite{garcia2017few} &  62.25\mypm0.65 & 44.28\mypm0.63 & 70.84\mypm0.65 & 52.53\mypm0.59 \\
	\hline
	Haar & 70.31\mypm0.67 & 46.11\mypm0.66 & 76.88\mypm0.63 & 57.72\mypm0.64 \\
	\hline
	Daubechies     & 70.03\mypm0.67 & 46.81\mypm0.63 & 77.17\mypm0.64 & 57.44\mypm0.66 \\
	\hline
	Coiflets   &  69.80\mypm0.67 & 47.94\mypm0.65 & 77.19\mypm0.65 & 57.51\mypm0.63 \\
	\hline
	Biorthogonal & 69.90\mypm0.69 & 47.75\mypm0.65 & 77.62\mypm0.65 & 58.03\mypm0.64 \\
	\hline
	Symlets & 68.69\mypm0.69 & 47.85\mypm0.67 & 77.97\mypm0.64 & 56.70\mypm0.64 \\
	\hline
	\end{tabular}
	\end{center}
	%\vspace{-0.1in}
	\caption{\textbf{The performance of wave-SAN with different wavelet transforms}. 5-way 5-shot results on FWT's benchmark are reported.}
\label{tab:waveletAlor}
%\vspace{-0.3in}
\end{table}

From the results, no matter which wavelet transform algorithm are used, our method has an obvious performance improvement when compared against the GNN baseline. The results indicate that our wave-SAN is robust to various wavelet transforms as long as they can serve as the tool for decomposing the high-frequency and low-frequency components.

\subsection{More Ablation Studies.}
Several important ablation studies are also considered, including the class selection strategy for dual episodes, how many wavelet transforms need to be performed before applying our StyleAug module, on which blocks to augment the styles, and the choices for the hyper-parameters $k_1$ and $k_2$. To answer these questions, we conduct experiments on the FWT's benchmark and the 5-way-1-shot results are reported in Table~\ref{tab:abla}.

	\begin{table*}[h] \small
		\begin{center}
			\begin{tabular} {c c c c c}
				\toprule
				\textbf{5-way-1-shot}  & \textbf{CUB} & \textbf{Cars} & \textbf{Places} & \textbf{Plantae} \\
				\hline
				& \multicolumn{4}{c}{same episode class set VS random episode class sets} \\
				\hline
				same class set 	& 49.16\mypm0.72 &	 32.47\mypm0.55	& 56.55\mypm0.81 & 37.79\mypm0.63 \\
				\hline  
				
				\textbf{random class sets (ours)}& \textbf{50.25\mypm0.74}  & \textbf{33.55\mypm0.61}& \textbf{57.75\mypm0.82} & \textbf{40.71\mypm0.66} \\
				\hline
				
				& \multicolumn{4}{c}{different number of wavelet transforms $J_{list}$} \\
				\hline
				\textbf{$\textbf{[1,1,1]}$ (ours)} & \textbf{50.25\mypm0.74}  & \textbf{33.55\mypm0.61}& 57.75\mypm0.82 & \textbf{40.71\mypm0.66} \\ 
				\hline
				$[2,2,2]$ & 50.24\mypm0.75  &	32.99\mypm0.61  & \textbf{58.27\mypm0.82}  & 39.43\mypm0.65  \\
				\hline
				$[3,3,3]$ & 48.60\mypm0.72  & 	33.35\mypm0.60  &  	56.51\mypm0.81  &   	39.27\mypm0.62   \\
				\hline
				$[1,2,3]$ & 49.74\mypm0.74  &	33.42\mypm0.59  &	57.36\mypm0.81  &	38.63\mypm0.62  \\
				\hline
				$[3,2,1]$ & 49.31\mypm0.73  &	33.14\mypm0.57  &	57.52\mypm0.82  &	39.01\mypm0.62  \\
				\hline  
			
				& \multicolumn{4}{c}{different blocks for StyleAug module} \\
				\hline
				%block 0 & 64.39\mypm0.79 & 46.91 \mypm0.71 &	32.56 \mypm0.56 &	55.61 \mypm0.81 &	37.55 \mypm0.61 \\
				%\hline
				block 1 & 47.59 \mypm0.71 &	32.50 \mypm0.57 &	54.47 \mypm0.80 &	37.15 \mypm0.60 \\
				\hline
				block 12 & 48.34 \mypm0.72 &	31.88 \mypm0.55 &	56.42 \mypm0.82 &	37.27 \mypm0.61 \\
				\hline
				\textbf{block 123 (ours)} &  \textbf{50.25\mypm0.74}  & \textbf{33.55\mypm0.61}& \textbf{57.75\mypm0.82} & \textbf{40.71\mypm0.66} \\
				\hline
				%block 0123 & 67.11\mypm0.81 & 48.55 \mypm0.70 & 	33.10 \mypm0.57 &	56.26 \mypm0.79 &	39.04 \mypm0.61 \\
				%\hline
				block 1234 & 44.37 \mypm0.65 &	34.11 \mypm0.58 &	54.74 \mypm0.79 &	37.29 \mypm0.61  \\
				\hline

				& \multicolumn{4}{c}{different choices for the loss function ($k_1:k_2$)} \\
				\hline
				1.0:0.0 & 47.93 \mypm0.72 &	31.26 \mypm0.51 &	55.40 \mypm0.82 &	37.20 \mypm0.58 \\
				\hline
				0.8:0.2 & 48.70 \mypm0.73 &	31.98 \mypm0.54 &	56.73 \mypm0.81 &	38.63 \mypm0.62 
				\\
				\hline
				0.5:0.5 & 48.03 \mypm0.71 & 	33.23 \mypm0.56 &	56.15 \mypm0.80 & 	38.84 \mypm0.61 \\
				\hline
				\textbf{0.2:0.8 (ours)} & \textbf{50.25\mypm0.74}  & \textbf{33.55\mypm0.61}& \textbf{57.75\mypm0.82} & \textbf{40.71\mypm0.66} \\
				\hline
				0.0:1.0  & 	49.44\mypm0.73 & 	32.77 \mypm0.59 & 57.39 \mypm0.81 &   39.34 \mypm0.62 \\
				\bottomrule
			\end{tabular}
		\end{center}
		\caption{\textbf{Ablation studies of our method.} The results for the 5-way-1-shot setting on FWT's benchmark are reported. }
		%\vspace{-0.3in}
		\label{tab:abla}
	\end{table*}

	\noindent\textbf{Sample the dual episodes from the same class set or randomly selected class sets?} We compare our randomly sampling strategy with that of sampling the two episodes of the same class set (``same class set''). We show that our strategy is obviously better than the ``same class set". This indicates that sampling the episodes from the same categories limits the diversity of the ``styles''. In contrast, exchanging the styles from different classes augments the source domain to a greater extent.
	
	\noindent\textbf{How many wavelet transforms should be performed?} In this paper, for each feature map extracted by the block of the feature extractor, we only apply once wavelet transform to decompose the low-frequency and high-frequency components. Factually, we can further apply the wavelet transform on the extracted low-frequency components. That is known as the multiple wavelet transforms. Notably, let $J$ denotes the number of wavelet transforms performed on a feature map, we have two characteristics: 1) no matter how much $J$, the wavelet transforms and the inverse wavelet transforms are totally reversible; 2) the larger $J$, the smaller the size of the low-frequency component, which also means that the extracted ``shape'' and ``styles'' are less. Therefore, we studied how the different combinations of $J$ on the first three blocks (denoted as $J_{list}$) will affect the final results. For example, $J_{list} = [1, 2, 3]$ means that we perform once, twice, and three wavelet transforms on the features of the first three blocks, respectively.
	We first observe that our choice $J_{list} = [1, 1, 1]$ is the best in most cases. Generally, comparing $J_{list} = [1, 1, 1]$, $J_{list} = [2, 2, 2]$, and $J_{list} = [3, 3, 3]$, when $J$ becomes larger, the corresponding performance decreases. This shows that more low-frequency components maintained in the feature map, more styles can be exchanged, thus resulting in better results. Therefore, for $J_{list} = [1, 2, 3]$ and $J_{list} = [3, 2, 1]$, their overall performances are also inferior to ours.

	\noindent\textbf{On which blocks to perform the style augmentation?} In our full wave-SAN model, the style augmentation is applied after block1, block2, and block3 (denoted as ``block 123''). We also perform the style augmentation on different blocks, resulting in the variants including ``block 1'', ``block 12'', ``block 123'' (ours), and ``block 1234''. For each block, the $J$ of wavelet transform is set as 1. Overall, comparing the ``block1'', ``block12'', and ``block123``, we find that as the number of blocks increases, the performance of the model shows a steady upward trend. This again indicates that the more styles exchanged, the more robust the model to various styles. We also observe that there is a performance degradation when comparing our ``block 123'' with ``block 1234''. This shows that the fourth block which is the top one of the feature extractor extracts more high-level semantic information. Exchanging such information may cause the semantic drift problem thus damaging the performance. These conclusions are generally consistent with that reported in the AdaIN~\cite{huang2017arbitrary}.
	
	\noindent\textbf{Different choices for the hyper-parameters of the final loss function.} Recall that our final loss is composed of $k_1 * \mathcal{L}_0$, $k_2 * \mathcal{L}_{aug}$, and $\mathcal{L}_{SSL}$. Thus, we tried different choices for the hyper-parameters $k_1$ and $k_2$ with different ratios. Results show that when we increase the $k_2$ from 0 to 0.8 gradually, the performance of the model is generally positively correlated. However, when we further increase the $k_2$ to 1.0, the performance is decreased. These phenomenons show that our style-augmented episodes play a very important role on the full model, while the initial episodes still contribute by showing the model what the real source images look like.

\section{Visualization Results}
\begin{figure*}[h]
	\centering
	\includegraphics[width=0.75\linewidth]{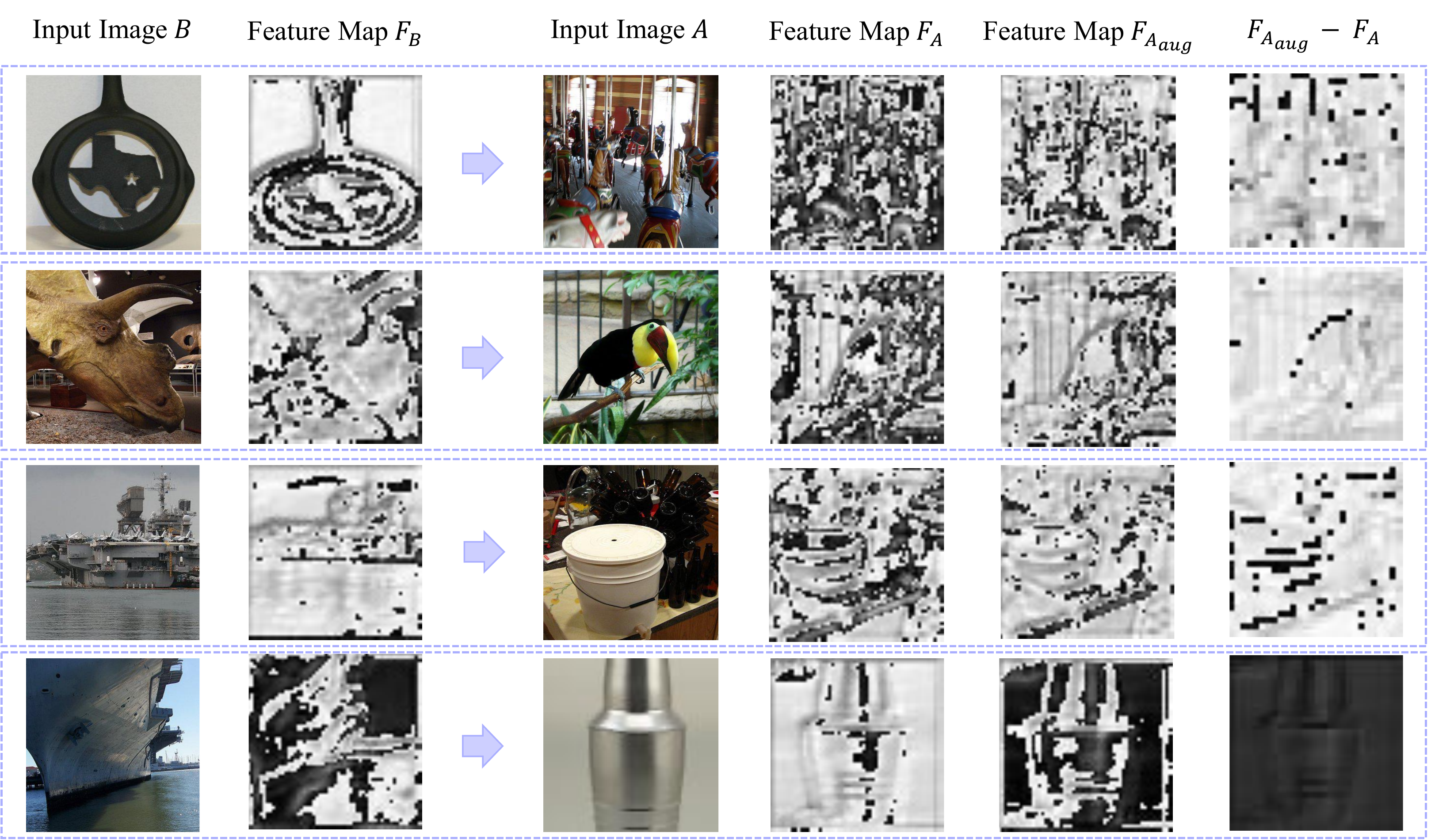}
	%\vspace{-0.1in}
	\caption{\textbf{The visualization of our style augmentation.} Given two images $A$ and $B$, the input images, the initial feature map $F_A$ and $F_B$, the style-augmented feature map $F_{A_{aug}}$ which is obtained by replacing the style of $F_A$ by that of $F_B$, and the results of $F_{A_{aug}} - F_A$ are displayed. Totally, five images randomly sampled from the mini-Imagenet are given.}
	\label{fig:styleAug} 
	%\vspace{-0.15in}
\end{figure*}

In order to show the impact of the style augmentation intuitively, given two input images $A$ and $B$, we augment the styles of $A$ by replacing styles of the low-frequency component of $A$ by that of $B$. As shown in Figure~\ref{fig:styleAug}, we visualize the input images $A$ (column 3) and $B$ (column 1), the initial feature map $F_A$ (column 4) and $F_B$ (column 2), the style-augmented feature map $F_{A_{aug}}$ (column 5). Besides, to better show the difference of $F_{A_{aug}}$ and $F_A$, the results of $F_{A_{aug}} - F_{aug}$ are demonstrated in the last column.  Concretely, the style augmentation is applied in the first block and the first channel of the feature map is visualized. 

We observe several points. 1) The basic visual information of the input images is well maintained in the feature maps $F_A$ and $F_B$, such as shape and edges. 2) Comparing the results of $F_A$ and $F_{A_{aug}}$, we notice that after performing the style augmentation, the key visual information of the image is not destroyed while the style is different from the original one. 3) The $F_{A_{aug}} - F_{A}$ better demonstrates the style diversity brought by our style augmentation method. Besides, by comparing $F_A$, $F_{A_{aug}}$, and $F_{A_{aug}} - F_A$ with $F_{B}$, we notice that the style of $F_A$ is indeed approaching that of $F_B$. This shows the effectiveness of our method.

\end{document}